\documentclass[11pt]{article}

\usepackage[T1]{fontenc}
\usepackage[letterpaper,margin=1in]{geometry}
\usepackage{amsmath,amssymb}
\usepackage{amsthm}
\usepackage{tgtermes}
\usepackage{newtxtext}
\usepackage{newtxmath}
\usepackage{bm}
\usepackage{graphicx}
\usepackage{subcaption}
\usepackage{tikz}
\usetikzlibrary{arrows.meta,positioning}
\usepackage{booktabs}
\usepackage{threeparttable}
\usepackage{algorithm}
\usepackage{placeins}
\usepackage{algpseudocode}
\usepackage{pgfplots}
\usepackage{xurl}
\usepackage{hyperref}
\usepackage[round,authoryear]{natbib}
\pgfplotsset{compat=1.18}

\bibpunct[, ]{(}{)}{,}{a}{}{,}%
\def\bibsep{\smallskipamount}%
\theoremstyle{remark}

\newtheorem{remark}{Remark}
\theoremstyle{definition}
\newtheorem{definition}{Definition}
\theoremstyle{plain}
\newtheorem{proposition}{Proposition}
\newtheorem{corollary}{Corollary}
\newcommand{\Fset}{\mathcal{F}}

\newcommand{\Iset}{\mathcal{I}}

\newcommand{\Tset}{\mathcal{T}}

\hypersetup{hidelinks}
\begin{document}
\raggedbottom

\title{Scaling Optimal Classification Trees via Adaptive Feature and Sample Reduction}
\author{%
Jiancheng Tu\\
\small Department of Computing, The Hong Kong Polytechnic University\\
\small \texttt{jiancheng.tu@connect.polyu.hk}
\and
Wenqi Fan\\
\small Department of Computing, The Hong Kong Polytechnic University\\
\small Department of Management and Marketing, The Hong Kong Polytechnic University\\
\small \texttt{wenqi.fan@polyu.edu.hk}
}
\date{}

\maketitle

\begin{abstract}
Dynamic programming for optimal classification trees becomes
computationally expensive as the numbers of  features and
training samples increase. We develop a joint feature- and
sample-space reduction framework based on \texttt{STreeD}.
\texttt{Weighted STreeD} merges duplicate records created after
projection onto a fixed candidate set into weighted representatives.
This reduces sample-dependent computation without changing
the fixed-candidate optimization problem.
\texttt{Adaptive STreeD}
repeatedly refines a bounded candidate set, retains features used by
the incumbent tree, rebuilds the weighted representation, and solves
the resulting reduced problems. Each certified
\texttt{Weighted STreeD} solution is optimal for its current
candidate set, while the outer feature search remains heuristic over
the full feature space.
Experiments on five data sets show that \texttt{Weighted STreeD}
achieves speedups of up to 121.41 times over standard
\texttt{STreeD}. \texttt{Adaptive STreeD} reduces runtime in matched
comparisons at depths 2--4 and continues to return feasible trees at
greater depths where full-feature methods are limited by time or
memory. Under the same computational budget, its predictive
performance remains comparable to the evaluated optimal classification tree
baselines and is higher in some  comparisons.
These results show how joint feature- and sample-space reduction can scale
dynamic-programming-based optimal-tree learning to more demanding
instances.
\end{abstract}

\noindent\textbf{Keywords:} Interpretable machine learning, optimal classification tree, dynamic programming

\section{Introduction}
\label{sec:introduction}

Classification trees are widely used when predictions must be
explained through explicit decision rules. This is particularly
relevant in high-stakes applications such as credit scoring, medical
prediction, and criminal justice, where decision makers may need to
inspect and justify individual predictions
\citep{rudin2019stop}. A shallow tree provides such transparency
through a short sequence of rules from the root node to a leaf.

Classical tree-learning methods such as CART, ID3, and C4.5 construct
trees greedily
\citep{breiman1984cart,quinlan1986induction,quinlan1993c45}.
They are computationally efficient, but local split decisions may not
lead to the best final tree. Optimal classification tree methods
instead optimize the complete tree. Existing approaches include
mixed-integer optimization
\citep{bertsimas2017optimal,verwer2019binary, alston2026mixed, liu2024optimal, blanquero2021optimal, blanquero2020sparsity}, dynamic programming
\citep{aglin2020dl85,demirovic2022murtree,lin2020gosdt,
vanderlinden2023streed}, SAT, and constraint programming
\citep{verhaeghe2020cp,shati2021sat}. Among these approaches,
dynamic programming exploits the recursive tree structure and has been
effective for shallow optimal-tree learning.

We build on \texttt{STreeD}, a general dynamic-programming framework
for optimal decision trees \citep{vanderlinden2023streed}. Its
computational cost depends strongly on the number of candidate
features and training samples. Large feature spaces increase the
number of states and split evaluations, while many states repeatedly
process training records. These costs become more severe after
preprocessing creates a high-dimensional binary feature space.

We exploit the interaction between feature and sample reduction.
Restricting the candidate set reduces split enumeration and can create
additional duplicate projected records, which are merged into weighted
representatives. Based on this idea, \texttt{Weighted STreeD} provides
a lossless reformulation for fixed candidate sets, and
\texttt{Adaptive STreeD} iteratively refines bounded candidate sets
while rebuilding the weighted representation.

The two methods have different guarantee scopes. A certified
\texttt{Weighted STreeD} solution is optimal for its fixed candidate
set. \texttt{Adaptive STreeD}, however, remains heuristic over the
full feature space. It recovers a full-feature optimum when a visited
candidate set contains the split features of such an optimal tree and
the corresponding reduced problem is solved to optimality. We also
derive complexity and conditional solution-quality results that
clarify these guarantees.

Experiments on five binarized data sets show that
\texttt{Weighted STreeD} achieves speedups of up to
\(121.41\times\) over standard \texttt{STreeD}.
\texttt{Adaptive STreeD} reduces runtime in matched comparisons at
depths 2--4 and continues to return feasible trees at greater depths
where full-feature methods are limited by time or memory. Under the
same computational budget, its predictive performance remains
comparable to the evaluated optimal-tree baselines.

The paper makes two methodological contributions. First, we develop a
joint feature- and sample-space reduction scheme in which candidate
restriction can increase the amount of lossless weighted aggregation.
Second, we develop an adaptive candidate-refinement method that
updates the reduced feature space while preserving features used by
the incumbent. Theoretical analysis characterizes the computational
reductions and guarantee scope, and computational experiments evaluate
the two mechanisms separately and jointly.

\section{Related Work}
\label{sec:related-work}

We organize the closest work by optimization, data and feature
reduction, and near-optimal search, emphasizing what each method
reduces and the scope of its guarantees.

\subsection{Optimal Classification Trees}

Optimal classification tree methods optimize the complete tree rather
than selecting splits greedily. Major exact approaches include
mixed-integer optimization, dynamic programming, SAT/MaxSAT,
constraint programming, and branch-and-bound.

Mixed-integer optimization models jointly determine tree structure,
split rules, and leaf predictions. The OCT formulation established
this approach for bounded-depth trees
\citep{bertsimas2017optimal}, followed by stronger formulations and
extensions to richer split rules, objectives, and constraints
\citep{verwer2019binary,aghaei2025strong,gunluk2021optimal,
subramanian2023multiway,ales2024new, donofrio2024margin}. These models are flexible, but
sample-indexed routing variables and large split spaces can limit
scalability. SAT, MaxSAT, and constraint-programming methods provide
alternative exact formulations based on logical constraints and
structured search
\citep{shati2021sat,hu2020maxsat,verhaeghe2020cp}. Column-generation
and path-based approaches further reduce the initial model by
generating rules or paths as needed
\citep{firat2020column,patel2024column,subramanian2023multiway}.

Dynamic-programming methods exploit the recursive structure of a
decision tree and reuse repeated subproblems through caching and
bounds. DL8.5 combines caching with branch-and-bound search
\citep{aglin2020dl85}, while MurTree introduces specialized bounds,
similarity-based pruning, and efficient shallow-tree routines
\citep{demirovic2022murtree}. OSDT and GOSDT use regularization and
strong pruning rules to learn sparse optimal trees
\citep{hu2019optimal,lin2020gosdt}. \texttt{STreeD} generalizes this line of
work by identifying conditions under which objectives and constraints
can be decomposed into independent subtree problems
\citep{vanderlinden2023streed}.
Recent alternatives include AND/OR search in MAPTree and AO*
search in Branches \citep{sullivan2024maptree,chaouki2025branches}.
Quant-BnB and ConTree instead specialize search for continuous
features \citep{mazumder2022quantbnb,brita2025continuous}.

\subsection{Data and Feature Reduction}
\label{subsec:data-feature-reduction}

Data-reduction methods use repeated or indistinguishable records to
reduce computation. CORELS and OSDT use equivalent-point arguments
to derive unavoidable prediction errors
\citep{angelino2018corels,hu2019optimal}. Similar ideas have been
used to strengthen lower bounds or reduce redundant observations in
optimal-tree search
\citep{zhang2023osrt,keegan2025acceleration,hua2022scalable}.

GOSDT provides a related dynamic-programming precedent. It stores
positive and negative empirical mass for each distinct row in a fixed
binary feature matrix \citep{lin2020gosdt}. WFlowOCT aggregates
duplicate feature--label records into weighted instances within a
flow-based mixed-integer model \citep{tu2026generalized}.
\texttt{Weighted STreeD} differs in that duplicate records are defined
after projection onto the current candidate set. Restricting the
candidate set can therefore create additional duplicate records,
which are merged into weighted representatives. The weighted
representation is rebuilt whenever the candidate set changes, coupling
feature-space reduction with sample-space reduction. Under the stated
separability and information-preservation conditions, the
transformation leaves the fixed-candidate optimization problem unchanged.

Feature reduction can be applied before optimization or during the
solution process. BinOCT reduces binary variables associated with
distinct feature values \citep{verwer2019binary}, while reference-guided methods use black-box
models to select thresholds, estimate tree size, or guide lower bounds
\citep{mctavish2022fast}. 
Other studies reduce the feature space more directly.
\citet{ruggieri2019complete} enumerates feature subsets for a fixed
greedy learner, while \citet{ing2024lad} and
\citet{eiben2023largedomain} study compact feature supports under
different structural assumptions. These methods reduce the search
space, but they do not generally guarantee that the selected features
contain those used by a full-feature optimal tree.
\texttt{Adaptive STreeD} follows a different
strategy: it solves a sequence of bounded candidate-set problems and
rebuilds the weighted data representation after each candidate update.

\subsection{Near-Optimal Classification Trees}

When proving full optimality is too expensive, several methods seek
high-quality trees within a fixed computational budget.
Limited-discrepancy search, Blossom, and anytime beam search retain the
full search space but prioritize strong incumbents before completing
the optimality proof
\citep{demirovic2023blossom,kiossou2025beam}.
Memory-constrained methods instead limit cache usage or modify state
processing
\citep{aglin2022memory}.
Other approaches reduce exact search by optimizing only part of the
tree: SPLIT solves upper subproblems exactly and constructs lower
levels greedily \citep{babbar2025split}, while SAT-based local improvement reoptimizes selected
subtrees of a heuristic solution
\citep{,schidler2021satlarge}.

\texttt{Adaptive STreeD} differs by restricting the candidate feature
space rather than the tree structure or search order. Each certified
inner solve is optimal for its current candidate set, while the outer
feature refinement remains heuristic over the full feature space. A
full-feature optimum is recovered only when a visited candidate set
contains the split features of such an optimal tree and the
corresponding reduced problem is solved to optimality.

\section{Weighted Unique-Data Dynamic Programming}
\label{sec:weighted-dp}

This section develops the fixed-candidate weighted reformulation of
\texttt{STreeD}. We first establish fixed-candidate equivalence between
the original and weighted representations for separable tasks that
satisfy the information-preservation condition. We then specialize to
the training-accuracy objective to present the explicit recursion and
analyze its time and space complexity. This specialization reflects
the paper's focus on scalability rather than a restriction of the
weighted reformulation. Proofs are given in
Appendix~\ref{app:proofs}.

\subsection{Problem Setting}
\label{subsec:problem-setting}

{
Let $\mathcal{I}=\{(\mathbf{x}_i,y_i,\eta_i)\}_{i=1}^{n}$ be a
training set, where $\mathbf{x}_i\in\{0,1\}^{p}$ is the
post-binarization feature vector, $y_i\in\mathcal{K}$ is the class
label drawn from the finite class set $\mathcal{K}$, and $\eta_i$
collects any additional record-level information
required by the task, such as class-specific costs, treatment and
outcome information, or group membership.

Let $\mathcal{F}=\{1,\ldots,p\}$ be the full post-binarization feature
set. Fix a candidate set $S\subseteq\mathcal{F}$ and let $q=|S|$.
A tree restricted to $S$ may use only features in $S$ as split
features; $h_T(\mathbf{x}_{iS})$ denotes the prediction of tree $T$
after projecting observation $i$ onto $S$.

For a tree $T$, let
$L_{\mathrm{mis}}(T)=\sum_{i=1}^{n}\mathbf{1}\{y_i\neq
h_T(\mathbf{x}_{iS})\}$ denote its training misclassification count.
The empirical misclassification error is $L_{\mathrm{mis}}(T)/n$, and
training accuracy is $1-L_{\mathrm{mis}}(T)/n$. Thus, maximizing
training accuracy is equivalent to minimizing $L_{\mathrm{mis}}(T)$.
The recursions below use this additive count.

\begin{definition}
\label{def:separable-task}
Following Definition~4.2 of \citet{vanderlinden2023streed}, an
optimization task is \emph{separable} if and only if the optimal
solution to any subtree can be determined independently of decision
variables outside that subtree and the branching decisions of its
parent nodes.
\end{definition}

With appropriate task states, \texttt{STreeD} supports ordinary and
cost-sensitive classification, prescriptive policy learning,
$F_1$-score and Matthews correlation coefficient optimization, and
gurop fairness objectives. The aggregation argument below applies
whenever the key or stored statistics retain all information used by
the task's dynamic program.

For the fixed candidate set $S$, define the aggregation key
\begin{equation}
\label{eq:aggregation-key}
g_i(S)=(\mathbf{x}_{iS},y_i,\eta_i).
\end{equation}
Observations with the same key form one group in the fixed-candidate
problem. The weighted unique data set $\mathcal{U}(S)$ contains one
representative $(\mathbf{z}_u,y_u,\eta_u,w_u)$ for each distinct key.
Here, $\mathbf{z}_u$ is the projected binary feature vector,
$y_u$ and $\eta_u$ retain the task-relevant information, and $w_u$
is the number of original observations represented by $u$. Let
\[
u_S = |\mathcal{U}(S)|
\]
denote the number of weighted unique records. Each representative must
retain all task-specific information and sufficient statistics required
by the dynamic program.

Let $\mathcal{T}(S,D,M)$ denote the class of classification trees with
depth at most $D$, at most $M$ internal split nodes, and split features
restricted to $S$. All results below use the same fixed $S$, the same
post-binarized representation, and the same feasible tree class. We use
$A$ and $B$ for states over original and weighted records, respectively,
and $d$ and $m$ for the remaining depth and node budget.
Appendix~\ref{app:notation} collects the notation.
}

\subsection{STreeD Dynamic Programming Baseline}
\label{subsec:streed-baseline}

For the training-accuracy objective, \texttt{STreeD} recursively solves the
two child subproblems created by a split \citep{vanderlinden2023streed}.
Its state $(A,d,m)$ records the observations reaching the node, the
remaining depth, and the remaining node budget. 
$\mathrm{Leaf}(A)$ is the smallest misclassification count attainable by a constant class prediction. For $f\in S$, let
\[
A_j(f)=\{i\in A:x_{if}=j\},\qquad j\in\{0,1\}.
\]
The cache avoids repeated solution of identical states. The simplified
recursion in Appendix~\ref{app:baseline-recursion} omits implementation
enhancements because it is used only to expose sample-dependent work.

Let $\mathcal{A}_{D,M}(S)$ be the set of distinct cached states
visited by the baseline recursion. For a state $(A,d,m)$, define
\[
n_A=|A|.
\]

We use the conservative path-state bound
\begin{equation}
\label{eq:path-state-bound}
H_D(q)
=
\sum_{\ell=0}^{\min\{D,q\}}
\binom{q}{\ell}2^\ell.
\end{equation}
When $D\le q$, treating $D$ as fixed and $q$ as the input variable,
\[
H_D(q)=O((2q)^D).
\]

The following proposition records the baseline time and space terms.
The bounds are intended for complexity accounting rather than as
tight models of a specific implementation.

\begin{proposition}
\label{prop:streed-time}
\label{prop:streed-space}
Consider the accuracy-based recursion above for labels in
the finite class set $\mathcal{K}$,
and suppose that the baseline implementation scans the records in every
visited state. The asymptotic input quantities are $n$, $q$, $D$, $M$,
and $|\mathcal{K}|$; $b$ denotes the machine-word size. Its running time is
\begin{equation}
\label{eq:streed-time-state}
T_{\mathrm{STreeD}}
=
O\!\left(
\sum_{(A,d,m)\in\mathcal{A}_{D,M}(S)}
\left[
(|\mathcal{K}|+q)n_A+qm
\right]
\right).
\end{equation}
A conservative worst-case bound is
\begin{equation}
\label{eq:streed-time-worst}
T_{\mathrm{STreeD}}
=
O\!\left(
(D+1)(M+1)H_D(q)
\left[
(|\mathcal{K}|+q)n+qM
\right]
\right).
\end{equation}

If each cached state is represented by a bitset over the $n$ original
records and one machine word stores $b$ bits, then the
sample-dependent cache-space requirement is
\begin{equation}
\label{eq:streed-space}
\mathrm{Space}_{\mathrm{STreeD}}
=
O\!\left(
|\mathcal{A}_{D,M}(S)|
\left(
\frac{n}{b}+1
\right)
\right).
\end{equation}
These expressions exclude storage for the input feature matrix,
returned trees, and auxiliary solver statistics.
\end{proposition}

The state-sum expressions separate work performed on the records from
split and budget enumeration. The worst-case form replaces the actual
visited-state collection by a conservative path bound, so it should be
read as an accounting bound rather than a prediction of realized
runtime. Pruning and state reuse can reduce realized work below this
bound.

\subsection{Weighted STreeD}
\label{subsec:weighted-streed}

\texttt{Weighted STreeD} applies the same fixed-candidate recursion to the
weighted unique data set $\mathcal{U}(S)$. The multiplicity weights
satisfy
\[
\sum_{u\in\mathcal{U}(S)}w_u=n.
\]

Let $\mathcal{E}$ denote the set of distinct task-specific values of
$\eta_i$. Because all projected features are binary,
\begin{equation}
\label{eq:unique-size-bound-revised}
u_S
=
|\mathcal{U}(S)|
\le
\min
\left\{
n,
|\mathcal{K}|\,|\mathcal{E}|\,2^{q}
\right\}.
\end{equation}
For the training-accuracy objective,
$|\mathcal{E}|=1$.

Equation~\eqref{eq:unique-size-bound-revised} is only an upper bound.
The actual value of $u_S$ depends on the patterns observed in the
data. When only a small fraction of the possible patterns occurs,
projection onto a small candidate set can create many duplicate
records. In such cases, $u_S$ can be much smaller than both the
original sample size and the combinatorial upper bound.
The main theoretical result is the equivalence between the original
and weighted fixed-candidate problems.

\begin{proposition}
\label{prop:objective-equivalence-revised}
Fix a candidate set $S$, a feasible tree class
$\mathcal{T}(S,D,M)$, and a separable optimization task. Suppose the
aggregation key retains all record-level information required by that
task. For each feasible tree $T$, let $\boldsymbol{\phi}_i(T)$ denote
record $i$'s contribution to the vector of sufficient statistics used
by the dynamic program. Assume that this contribution depends only on
$h_T(\mathbf{x}_{iS})$, $y_i$, and $\eta_i$, and that the task value
and feasibility of $T$ are determined by the sum of these contributions
together with any tree-only terms. If $\boldsymbol{\phi}_u(T)$ denotes
the corresponding contribution for weighted record $u$, then
\begin{equation}
\label{eq:weighted-objective-equivalence}
\sum_{i=1}^{n}\boldsymbol{\phi}_i(T)
=
\sum_{u\in\mathcal{U}(S)}w_u\boldsymbol{\phi}_u(T).
\end{equation}
The original and weighted representations therefore give every
feasible tree the same task statistics and the same value under the
possibly nonlinear task objective computed from those statistics.
Because the feasible tree class and tree-only terms are unchanged, the two
fixed-candidate problems also have the same optimal value and the same
set of optimal trees.
\end{proposition}

\begin{remark}
\label{rem:weighted-general}
The weighted representation supports the same separable tasks as
\texttt{STreeD} whenever the aggregation key or stored sufficient
statistics retain the information required by the task. What must be
retained is task dependent. For the training-accuracy objective,
projected features and class labels suffice. For the nonlinear but separable
$F_1$-score objective, the weighted counts must reproduce the class-specific
confusion-matrix statistics used by the task state.
\end{remark}

We now specialize the weighted recursion and its complexity analysis
to the training-accuracy objective.
For this weighted recursion, let
\[
B\subseteq\mathcal{U}(S)
\]
be the weighted records reaching a node. The best weighted leaf cost is
\begin{equation}
\label{eq:weighted-best-leaf}
\begin{aligned}
\mathrm{Leaf}(B)
&=
\min_{k\in\mathcal{K}}
\sum_{u\in B}w_u\mathbf{1}\{y_u\neq k\}
\\
&=
\sum_{u\in B}w_u
-
\max_{k\in\mathcal{K}}
\sum_{u\in B}
w_u\mathbf{1}\{y_u=k\}.
\end{aligned}
\end{equation}

For a split feature $f\in S$, define
\[
B_0(f)
=
\{u\in B:z_{uf}=0\},
\qquad
B_1(f)
=
\{u\in B:z_{uf}=1\}.
\]

Let $V(B,d,m)$ be the minimum weighted misclassification count for
state $(B,d,m)$. The recursion is
\begin{equation}
\label{eq:weighted-dp-recursion-revised}
V(B,d,m)
=
\begin{cases}
\mathrm{Leaf}(B),
&
d=0 \text{ or } m=0,
\\[3pt]
\min
\left\{
\mathrm{Leaf}(B),
\Phi(B,d,m)
\right\},
&
\text{otherwise},
\end{cases}
\end{equation}
where
\begin{equation}
\label{eq:weighted-split-subproblem}
\Phi(B,d,m)
=
\min_{\substack{f\in S:\\
B_0(f)\neq\emptyset,\;
B_1(f)\neq\emptyset}}
\;
\min_{\substack{
m_0+m_1=m-1\\
m_0,m_1\ge 0}}
\left[
V(B_0(f),d-1,m_0)
+
V(B_1(f),d-1,m_1)
\right].
\end{equation}

The weighted and unweighted recursions have the same structure. Their
only difference is the representation of the records and the use of
weights in the sufficient statistics.

\begin{corollary}
\label{prop:weighted-recursion-validity}
For the training-accuracy objective with labels in the finite class set
$\mathcal{K}$, suppose the weighted and unweighted
recursions use the same fixed candidate set $S$, depth limit $D$,
internal-node budget $M$, and pruning rules. If those rules are computed
from aggregation-preserving weighted class counts, then
\eqref{eq:weighted-dp-recursion-revised}--%
\eqref{eq:weighted-split-subproblem} returns the same optimal
misclassification count as the corresponding recursion on the original
data.
\end{corollary}

Let $\mathcal{B}_{D,M}(S)$ be the set of distinct weighted states
visited by \texttt{Weighted STreeD}. For a weighted state $(B,d,m)$, define
$u_B=|B|$.

The following proposition gives the weighted time and space terms.

\begin{proposition}
\label{prop:weighted-streed-time}
\label{prop:weighted-streed-space}
Consider the weighted misclassification-count recursion
above for labels in the finite class set $\mathcal{K}$, and suppose
that \texttt{Weighted STreeD} uses the same direct
state-scanning implementation as the baseline. The asymptotic input
quantities are $u_S$, $q$, $D$, $M$, and $|\mathcal{K}|$; $b$ denotes
the machine-word size. Its running time is
\begin{equation}
\label{eq:weighted-streed-time-state}
T_{\mathrm{W\text{-}STreeD}}
=
O\!\left(
\sum_{(B,d,m)\in\mathcal{B}_{D,M}(S)}
\left[
(|\mathcal{K}|+q)u_B+qm
\right]
\right).
\end{equation}
Using the path-state bound in
\eqref{eq:path-state-bound}, a conservative worst-case bound is
\begin{equation}
\label{eq:weighted-streed-time-worst}
T_{\mathrm{W\text{-}STreeD}}
=
O\!\left(
(D+1)(M+1)H_D(q)
\left[
(|\mathcal{K}|+q)u_S+qM
\right]
\right).
\end{equation}

If each weighted state is represented by a bitset over the $u_S$
weighted unique records, then the sample-dependent cache-space
requirement is
\begin{equation}
\label{eq:weighted-streed-space}
\mathrm{Space}_{\mathrm{W\text{-}STreeD}}
=
O\!\left(
|\mathcal{B}_{D,M}(S)|
\left(
\frac{u_S}{b}+1
\right)
\right).
\end{equation}
These expressions exclude storage for the weighted input table,
returned trees, and auxiliary solver statistics.
\end{proposition}

Relative to the baseline bounds, the sample-universe terms replace
$n$ and the state sizes $n_A$ by $u_S$ and $u_B$. Candidate-set size,
depth, and node-budget factors remain. The weighted representation can
reduce record-processing work and bitset length without eliminating the
combinatorial search over feasible trees. The exact reduction in
sample-scanning work depends on the compression achieved within all
visited states, not only at the root state.

\begin{proposition}
\label{prop:sample-dependent-reduction}
For the training-accuracy objective with labels in the
finite class set $\mathcal{K}$, assume that the baseline and weighted
recursions visit corresponding logical states under the
same candidate set, depth limit, internal-node budget, search order,
and aggregation-preserving pruning rules. Then the reduction ratio in
the sample-scanning term is
\begin{equation}
\label{eq:time-reduction-exact}
R_T^{\mathrm{scan}}
=
\frac{
\displaystyle
\sum_{(A,d,m)\in\mathcal{A}_{D,M}(S)}
n_A
}{
\displaystyle
\sum_{(B,d,m)\in\mathcal{B}_{D,M}(S)}
u_B
}.
\end{equation}

If the state-level compression ratios are close to the root-level
compression ratio, then
\begin{equation}
\label{eq:complexity-reduction-summary}
R_T^{\mathrm{scan}}
\approx
\frac{n}{u_S}.
\end{equation}

For bitset-based cache representations and corresponding logical
states, the per-state sample-dependent storage ratio is
\begin{equation}
\label{eq:space-reduction-exact}
R_S^{\mathrm{bitset}}
=
\frac{n/b+1}{u_S/b+1}.
\end{equation}
When both $n/b$ and $u_S/b$ are large,
\begin{equation}
R_S^{\mathrm{bitset}}
\approx
\frac{n}{u_S}.
\end{equation}
\end{proposition}

The ratio \(n/u_S\) approximates sample-scanning and
bitset-storage reductions only when similar compression persists
across visited states. It is not a wall-clock speedup because feature
enumeration, budget allocation, caching, pruning, and fixed solver
overhead remain. The fixed-candidate ablation in
Section~\ref{subsec:weighted-ablation} evaluates the realized effect.
When \(S\) changes, projection and aggregation are repeated, so the
equivalence and complexity results apply separately to each candidate
set.

\section{Adaptive STreeD}
\label{sec:candidate-refinement}

Section~\ref{sec:weighted-dp} gives a lossless weighted reformulation for a fixed candidate feature set. \texttt{Adaptive STreeD} applies
this reformulation to a sequence of reduced problems with bounded
candidate sets in large binarized feature spaces. The initial candidate
set consists of the \(K\) most important features selected by a random
forest. Later iterations retain the features used by the current
incumbent and use repeated CART fits to propose new candidates. After
each update, the data are projected onto the active feature set, which
can create additional duplicate records. These records are merged into
weighted representatives, and the resulting problem is solved by
\texttt{Weighted STreeD}. Random forests and CART are used only for
feature proposal. A certified inner solve is optimal for its current
candidate set under the assumptions of Section~\ref{sec:weighted-dp}, whereas the outer candidate refinement
remains heuristic over the full feature space.

\subsection{Candidate-Refinement Algorithm}
\label{subsec:candidate-algorithm}

Algorithm~\ref{alg:candidate-refinement} summarizes the complete
\texttt{Adaptive STreeD} procedure.
Let \(p=|\Fset|\), and let the target tree have maximum depth
\(D\) and node budget \(M\) (\(M=2^D-1\) when unrestricted).
The training score is \(Q(T;\Iset)\in[0,1]\); the experiments use
accuracy. Each active set \(E_t\) contains at most \(K\) features.

\begin{algorithm}[t]
\caption{Adaptive STreeD}
\label{alg:candidate-refinement}
\footnotesize
\begin{algorithmic}[1]

\Require Data \(\Iset\), features \(\Fset\), depth \(D\), node budget
\(M\), score \(Q\), capacity \(K\), time limits
\((T_{\max},\bar{\tau})\), iteration limit \(N_{\max}\), patience
\(P_{\max}\), switch threshold \(P_{\mathrm{switch}}\), tolerance
\(\epsilon\)

\Ensure Best feasible tree \(T^{\mathrm{inc}}\)

\State \(E\gets\) top-\(K\) features ranked by a random forest
\State \(T^{\mathrm{inc}}\gets\) best constant-leaf tree;
\(Q^{\mathrm{inc}}\gets Q(T^{\mathrm{inc}};\Iset)\)
\State \(C\gets\emptyset\), \(R\gets\emptyset\), \(P\gets E\),
\(r\gets0\)

\For{\(t=1,\ldots,N_{\max}\)}

    \If{\(\operatorname{elapsed}\ge T_{\max}\) or
    \(Q^{\mathrm{inc}}\ge1-\epsilon\)}
        \State \textbf{break}
    \EndIf

    \If{\(t>1\)}
        \If{\(|C|\ge K\)}
            \State \textbf{break}
        \EndIf

        \State \(G\gets R\) if \(r<P_{\mathrm{switch}}\);
        otherwise \(G\gets R\cup P\)
        \State \(N\gets\) at most \(K-|C|\) top features from
        repeated CART fits on \(\Fset\setminus G\)

        \If{\(N=\emptyset\)}
            \State \textbf{break}
        \EndIf

        \State \(E\gets C\cup N\); \(P\gets P\cup N\)
    \EndIf

    \State \(\tau\gets
    \min\{\bar{\tau},\,T_{\max}-\operatorname{elapsed}\}\)
    \State \(\mathcal{U}(E)\gets\) weighted unique data obtained from
    the projection of \(\Iset\) onto \(E\)
    \State \(T\gets
    \mathrm{WeightedSTreeD}(\mathcal{U}(E),E,D,M,\tau)\)

    \If{\(T\) is feasible and
    \(Q(T;\Iset)>Q^{\mathrm{inc}}+\epsilon\)}
        \State \(T^{\mathrm{inc}}\gets T\);
        \(Q^{\mathrm{inc}}\gets Q(T;\Iset)\)
        \State \(C\gets\operatorname{supp}(T)\);
        \(R\gets R\cup C\); \(r\gets0\)
    \Else
        \State \(r\gets r+1\)
    \EndIf

    \If{\(r\ge P_{\max}\)}
        \State \textbf{break}
    \EndIf

\EndFor

\State \Return \(T^{\mathrm{inc}}\)

\end{algorithmic}
\end{algorithm}

A depth-\(D\) binary tree has at most \(2^D-1\) internal nodes and
therefore cannot use more than \(2^D-1\) distinct split features.
The capacity \(K\) exploits this structural sparsity. A smaller value
reduces split enumeration and usually increases compression after
projection; a larger value supplies broader feature coverage but
makes the exact reduced problem harder. The sensitivity experiment in
Section~\ref{subsec:sen_can_fea} evaluates this tradeoff directly.

The initial candidate set contains the top-\(K\) features ranked by
random forest, and the incumbent is initialized as the best constant
leaf. At later iterations, the incumbent support \(C\) is retained and
at most \(K-|C|\) additional features are proposed using repeated
multi-level CART fits, which can identify features that become useful
after earlier splits. The archives \(R\) and \(P\) record features used
by accepted incumbents and previously proposed features, respectively.
After \(P_{\mathrm{switch}}\) nonimproving iterations, the proposal
step also excludes \(P\), encouraging exploration of new feature
blocks. CART is used only for feature proposal and provides no
optimality guarantee.

The incumbent support is never removed. If \(|C|=K\), no new feature
can be added while retaining the incumbent, and the procedure stops;
otherwise, the proposal block is truncated so that
\(E_t=C_{t-1}\cup N_t\) satisfies \(|E_t|\le K\). For each active
candidate set, the data are projected, aggregated, and solved by
\texttt{Weighted STreeD}. Let \(T_{\max}\) be the total time limit and
\(\bar{\tau}\) the maximum time for one inner solve. Provided that
\(\operatorname{elapsed}<T_{\max}\), the time assigned to iteration
\(t\) is
\begin{equation}
\label{eq:inner-time-limit}
\tau_t=
\min\left\{
\bar{\tau},
T_{\max}-\operatorname{elapsed}
\right\}.
\end{equation}

Only an improvement greater than \(\epsilon\) replaces the incumbent;
failed or nonimproving inner solves leave it unchanged. Hence, the
incumbent-score sequence is nondecreasing, and an interrupted inner
solve cannot discard the best feasible tree found previously. A
certified inner solution is optimal for the current candidate set,
whereas an uncertified feasible solution is only an incumbent for that
reduced problem. Retaining the incumbent support keeps the current
tree feasible in subsequent reduced problems as long as its support
fits within the candidate capacity.

\subsection{Joint Feature- and Sample-Space Reduction}
\label{subsec:joint-reduction}

At iteration \(t\), let \(q_t=|E_t|\le K\) and
\(u_t=|\mathcal{U}(E_t)|\). Feature restriction replaces the full
dimension \(p\) by \(q_t\), while weighted aggregation replaces the
original \(n\) records by \(u_t\) weighted records in the
sample-dependent terms of the inner dynamic program.
For binarized features and labels in $\mathcal{K}$,
\begin{equation}
\label{eq:adaptive-unique-bound}
u_t
\le
\min\left\{
n,
|\mathcal{K}|2^{q_t}
\right\}.
\end{equation}
This is an upper bound. The actual value of \(u_t\) depends on the
feature--label patterns observed after projection and may be much
smaller.
The following proposition gives the per-iteration complexity under
the model developed in Section~\ref{sec:weighted-dp}.

\begin{proposition}
\label{prop:bounded-subproblem}
Consider classification with labels in the finite class set
$\mathcal{K}$, binarized candidate features, and the training-accuracy
objective. Suppose that iteration \(t\) uses
candidate set \(E_t\), with \(q_t=|E_t|\le K\), and aggregation key
\((\mathbf{x}_{iE_t},y_i)\). Under the direct state-scanning model of
Proposition~\ref{prop:weighted-streed-time}, the running time of the
inner \texttt{Weighted STreeD} solve is
\begin{equation}
\label{eq:adaptive-inner-complexity}
O\!\left(
(D+1)(M+1)H_D(q_t)
\left[
(|\mathcal{K}|+q_t)u_t+q_tM
\right]
\right),
\end{equation}
where \(u_t\le\min\{n,|\mathcal{K}|2^{q_t}\}\).
The asymptotic input quantities are $n$, $q_t$, $D$, $M$,
$K$, and $|\mathcal{K}|$.
Since \(q_t\le K\), a uniform per-iteration bound is
\begin{equation}
\label{eq:bounded-core}
O\!\left(
(D+1)(M+1)H_D(K)
\left[
(|\mathcal{K}|+K)
\min\{n,|\mathcal{K}|2^K\}
+
KM
\right]
\right).
\end{equation}
\end{proposition}

The capacity \(K\) reduces two sources of computational cost. First,
each dynamic-programming state considers at most \(K\) candidate
features instead of all \(p\) features. Second, projection onto a
smaller feature set can create fewer distinct records.
\texttt{Weighted STreeD} therefore processes fewer weighted records.
Once the active candidate set is constructed, the complexity of the
inner dynamic program depends on \(K\) rather than the full feature
dimension \(p\). The complete \texttt{Adaptive STreeD} procedure,
however, still depends on the original data because feature proposal,
projection, and aggregation operate on the full input.

For \(N\) completed iterations, let \(C_{\mathrm{RF}}(n,p)\) denote
the cost of the initial random-forest ranking and
\(C_{\mathrm{CART},t}(n,p)\) the cost of the CART proposal at
iteration \(t\). Assuming expected constant-time hash-table
operations, projection and aggregation require
\(O(n\sum_{t=1}^{N}q_t)\) expected work. The total computational cost
is therefore
\begin{equation}
\label{eq:adaptive-total-cost}
C_{\mathrm{RF}}(n,p)
+
\sum_{t=2}^{N} C_{\mathrm{CART},t}(n,p)
+
O\!\left(
n\sum_{t=1}^{N} q_t
\right)
+
\sum_{t=1}^{N} T_{\mathrm{DP},t},
\end{equation}
where \(T_{\mathrm{DP},t}\) is bounded by
\eqref{eq:adaptive-inner-complexity}.
Here $N$, $n$, $p$, and the candidate sizes $q_t$ are input
quantities. Equation~\eqref{eq:adaptive-total-cost} accounts for the
entire pipeline. In contrast, $T_{\mathrm{DP},t}$ covers only the exact
dynamic-programming solve at iteration $t$.

\begin{figure}[t]
\centering
\resizebox{\textwidth}{!}{%
\begin{tikzpicture}[
  font=\small,
  node distance=4.5mm and 5mm,
  box/.style={draw=black, rounded corners=1.2pt, align=left, inner sep=5pt,
    minimum height=11mm, text width=0.175\textwidth, fill=white},
  effect/.style={draw=black, dashed, align=center, inner sep=4pt,
    text width=0.17\textwidth, fill=black!4},
  arrow/.style={-{Latex[length=2.2mm]}, line width=0.6pt},
  loop/.style={-{Latex[length=2.2mm]}, line width=0.6pt, rounded corners=3pt}
]
\node[box] (data) {\textbf{Original data}\\
  \(n\) observations\\
  \(p\) binary features};
\node[box, right=of data] (proposal) {\textbf{1. Feature proposal}\\
  \(p\rightarrow q,\ q\le K\le p\)\\
  {\scriptsize cost \(C_{\mathrm{RF}}(n,p)\) if \(t=1\); }\\[-2pt]
  {\scriptsize cost \(C_{\mathrm{CART},t}(n,p)\) if \(t\ge 2\)}};
\node[box, right=of proposal] (projection) {\textbf{2. Projection}\\
  \(X\rightarrow X_S\)\\
  expected cost \(O(nq)\)};
\node[box, right=of projection] (aggregation) {\textbf{3. Aggregation}\\
  \(n\rightarrow u_S\)\\
  expected cost \(O(nq)\)\\
  \(u_S\le\min\{n,|\mathcal K|2^q\}\)};
\node[box, right=of aggregation, text width=0.205\textwidth] (solve)
  {\textbf{4. Weighted STreeD solve}\\
  input \(u_S\times q\)\\[-2pt]
  {\scriptsize\(\displaystyle O\!\left((D+1)(M+1)H_D(q)\right.\)}\\[-2pt]
  {\scriptsize\(\displaystyle \left.{}\times[(|\mathcal K|+q)u_S+qM]\right)\)}};

\draw[arrow] (data) -- (proposal);
\draw[arrow] (proposal) -- (projection);
\draw[arrow] (projection) -- (aggregation);
\draw[arrow] (aggregation) -- (solve);

\node[effect, below=7mm of proposal] (featureeffect)
  {\textbf{Feature-space reduction}\\
   \(H_D(p)\rightarrow H_D(q)\)\\
   \(H_D(q)=O((2q)^D)\)};
\node[effect, below=7mm of aggregation] (sampleeffect)
  {\textbf{Sample-space reduction}\\
   \(n\rightarrow u_S\)};
\draw[arrow] (proposal) -- (featureeffect);
\draw[arrow] (aggregation) -- (sampleeffect);

\node[box, below=7mm of solve, text width=0.205\textwidth] (update)
  {\textbf{5. Incumbent update}\\
   retain useful split features;
   refine the next candidate set};
\draw[arrow] (solve) -- (update);
\draw[loop] (update.east) -- ++(6mm,0)
  |- ([yshift=7mm]proposal.north) -- (proposal.north);
\end{tikzpicture}
}
\caption{One Adaptive STreeD refinement cycle. Candidate restriction
reduces \(p\) features to \(q_t\), and aggregation maps \(n\) projected
records to \(u_t\) weighted unique records before the inner solve.}
\label{fig:workflow}
\end{figure}
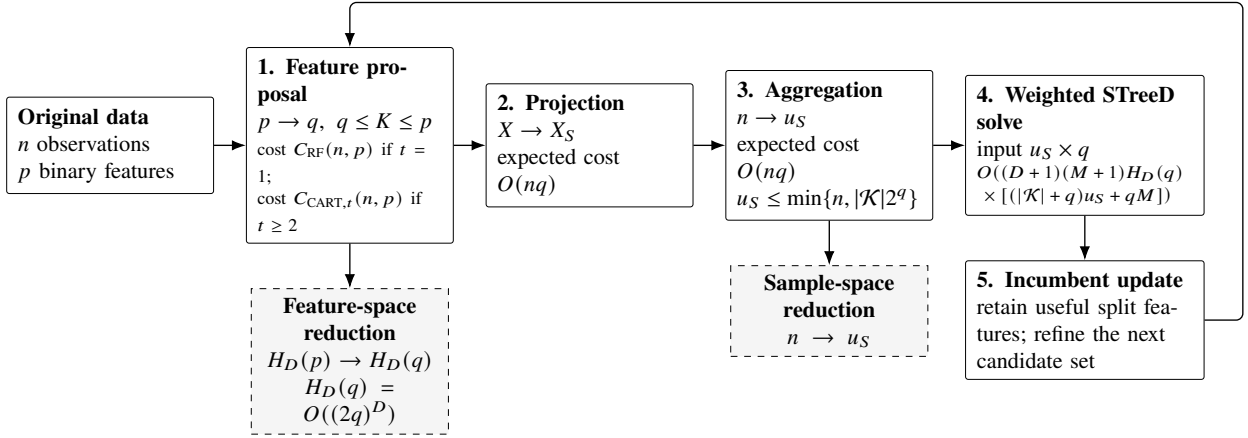

Equation~\eqref{eq:adaptive-total-cost} shows that the complete
procedure still depends on the original sample size \(n\) and feature
dimension \(p\), even though each inner dynamic-programming problem
operates on a reduced representation. Accordingly, \(n/u_t\) and
\(H_D(p)/H_D(q_t)\) quantify the reductions in the sample- and
feature-dependent components of the dynamic program, respectively.
They should not be interpreted as guarantees on total wall-clock
speedup.
Figure~\ref{fig:workflow} summarizes one refinement cycle of \texttt{Adaptive STreeD}.

\subsection{Conditional Solution-Quality Analysis}
\label{subsec:probabilistic}

We examine how the main parameters of \texttt{Adaptive STreeD}
influence solution quality, with particular attention to training
accuracy. Recovering a full-feature optimum requires that some visited
candidate set contain the split features of an optimal tree and that
the corresponding reduced problem be solved to certified optimality.
This section relates this event to candidate capacity, refinement
opportunities, and the quality of the retained incumbent.

Let
\begin{equation}
\label{eq:full-feature-optimum}
Q_D^\star
=
\max_{T\in\Tset(\Fset,D,M)}
Q(T;\Iset)
>0.
\end{equation}
Choose a full-feature optimal tree \(T_D^\star\), and define
\[
S_D^\star=\operatorname{supp}(T_D^\star),
\qquad
K^\star=|S_D^\star|.
\]
Since a feasible tree has depth at most \(D\) and at most \(M\)
internal nodes,
\[
K^\star\leq \min\{M,2^D-1\}.
\]
We assume \(K^\star\leq K\), so that an active candidate set can
contain the selected optimal support.

For iteration \(t\), define
\[
\mathcal{A}_t
=
\{S_D^\star\subseteq E_t\},
\]
and
\[
\mathcal{B}_{t-1}
=
\bigcap_{s=1}^{t-1}\mathcal{A}_s^c,
\]
with \(\mathcal{B}_0\) equal to the whole sample space. Let
\(\mathcal{H}_{t-1}\) denote the search history before iteration \(t\).
Suppose deterministic constants
\(\underline{\rho}_t\in[0,1]\) satisfy
\begin{equation}
\label{eq:conditional-coverage}
\Pr\!\left(
\mathcal{A}_t\mid\mathcal{H}_{t-1}
\right)
\geq
\underline{\rho}_t
\qquad
\text{almost surely on }\mathcal{B}_{t-1}.
\end{equation}

\begin{proposition}
\label{prop:expected-approximation}
Assume that \eqref{eq:conditional-coverage} holds for
\(t=1,\ldots,N\), and that whenever \(\mathcal{A}_t\) occurs, the
corresponding reduced problem is solved to certified optimality.
Suppose further that the retained incumbent has approximation ratio
at least \(r_0\in[0,1]\). If \(Q_{A,N}\) denotes the final incumbent
score after \(N\) completed iterations, then
\begin{equation}
\label{eq:expected-approximation}
\mathbb{E}\!\left[
\frac{Q_{A,N}}{Q_D^\star}
\right]
\geq
1-
(1-r_0)
\prod_{t=1}^{N}
(1-\underline{\rho}_t).
\end{equation}
\end{proposition}

Proposition~\ref{prop:expected-approximation} highlights three factors:
the baseline ratio \(r_0\), the number of completed iterations \(N\),
and the conditional support-coverage bounds
\(\underline{\rho}_t\). Larger values of any of these quantities
improve the bound. The \(\underline{\rho}_t\) are theoretical lower
bounds; random-forest and CART importance scores are used only to rank
candidate features and are not estimates of these probabilities.

For training accuracy, let
\[
\pi_{\max}
=
\max_{k\in\mathcal{K}}
\frac{1}{n}
\sum_{i=1}^{n}
\mathbf{1}\{y_i=k\}
\]
denote the majority-class proportion. Since
Algorithm~\ref{alg:candidate-refinement} initializes the incumbent
with the majority-class leaf and \(Q_D^\star\leq 1\), the initial
approximation ratio is at least \(\pi_{\max}\).

\begin{corollary}
\label{cor:accuracy-approximation}
For training accuracy, under the assumptions of
Proposition~\ref{prop:expected-approximation},
\begin{equation}
\label{eq:accuracy-approximation}
\mathbb{E}\!\left[
\frac{Q_{A,N}}{Q_D^\star}
\right]
\geq
1-
(1-\pi_{\max})
\prod_{t=1}^{N}
(1-\underline{\rho}_t).
\end{equation}
\end{corollary}

The candidate capacity \(K\) affects both feature coverage and the
difficulty of the reduced problem. The condition \(K^\star\leq K\)
is necessary for an active set to contain the selected optimal
support. Increasing \(K\) may improve coverage, but it also enlarges
the reduced problem and can make certification more difficult. The
random-forest initialization and CART-guided refinement determine
which features enter the candidate sets and therefore influence the
coverage probabilities \(\underline{\rho}_t\).

The remaining parameters mainly control how many reduced problems can
be explored and certified. The iteration and patience limits affect
the number of completed iterations \(N\). A larger inner time limit
\(\bar{\tau}\) gives each reduced problem more time to reach certified
optimality, whereas the total budget \(T_{\max}\) limits both the
number and duration of inner solves. Tree depth \(D\) and node budget
\(M\) affect the possible support size through
\(K^\star\leq\min\{M,2^D-1\}\) and also determine the difficulty of
the dynamic program. Under a fixed computational budget, training
accuracy therefore depends on the balance between feature coverage,
refinement opportunities, and the ability to certify the resulting
reduced problems.

\section{Computational Experiments}
\label{sec:experiments}

The computational study addresses three questions:
\begin{enumerate}
    \item[\textbf{Q1.}]
    Does \texttt{Adaptive STreeD} improve the scalability of optimal
    classification tree learning while maintaining predictive
    performance as the maximum tree depth increases?

    \item[\textbf{Q2.}]
    When the candidate feature set is fixed, to what extent does
    weighted unique-data aggregation reduce the computational cost
    of \texttt{STreeD}?

    \item[\textbf{Q3.}]
    How does the candidate-set capacity $K$ affect the tradeoff
    between computational efficiency and predictive performance?
\end{enumerate}

We conduct three experiments. The overall comparison evaluates
\texttt{Adaptive STreeD} against CART and two full-feature optimal-tree solvers
across different tree depths. The fixed-candidate ablation isolates
the computational effect of \texttt{Weighted STreeD}. The candidate-capacity
study examines the sensitivity of \texttt{Adaptive STreeD} to $K$.
Appendix~\ref{app:binning-sensitivity-weighted-unique} separately
examines the preprocessing resolution used to construct the binary
candidate universe.

\subsection{Experimental Setup}
\label{subsec:experimental-setup}

\textbf{Data and preprocessing.}
We use five public tabular data sets: COMPAS
\citep{angwin2016machine,propublica2016compas}, Diabetes 130-US
Hospitals \citep{strack2014impact,clore2014diabetes}, Give Me Some
Credit \citep{kaggle2011giveme}, FICO HELOC
\citep{fico2018heloc,arya2020aix360}, and Capital One Transactions
\citep{capitalone2018transactions,kagglefrauddetection}. We use five
independent 80/20 train--test splits. 
Continuous features are quantile
binned into at most 100 intervals and represented by cumulative binary
indicators; categorical features are one-hot encoded. The 100-bin cap
is used as a conservative preprocessing choice to retain more
threshold information before optimization, rather than to minimize
runtime. Appendix~\ref{app:binning-sensitivity-weighted-unique}
examines the sensitivity to this choice, and
Table~\ref{tab:dataset-summary} reports the resulting Bin100
dimensions.

\begin{table}[t]
\centering
\caption{Data sets used in the computational experiments}
\label{tab:dataset-summary}
\begin{threeparttable}
\small
\begin{tabular}{lrrrr}
\toprule
Data set
& Samples
& Original features
& Binary features
& Classes \\
\midrule
COMPAS
& 12,381
& 22
& 230
& 2 \\

Diabetic
& 101,766
& 47
& 544
& 3 \\

Give Me Some Credit
& 150,000
& 10
& 375
& 2 \\

FICO
& 10,459
& 23
& 758
& 2 \\

Transactions
& 786,363
& 113
& 653
& 2 \\
\bottomrule
\end{tabular}
\end{threeparttable}
\end{table}

\textbf{Method settings.}
The overall comparison uses \(K=20\). Unless otherwise stated,
\texttt{Adaptive STreeD} uses a total time limit of
\(T_{\max}=600\) seconds, a maximum inner-solve time of
\(\bar{\tau}=100\) seconds, \(N_{\max}=300\) outer iterations,
\(P_{\max}=3\), \(P_{\mathrm{switch}}=2\), and an improvement tolerance of
\(\epsilon=10^{-9}\). The CART proposal procedure uses depth 5 and
20 repeated fits per iteration. The sensitivity experiment varies \(K\) while
holding the remaining parameters fixed.
\texttt{STreeD}, \texttt{Weighted STreeD}, and DL8.5 optimize
training accuracy with the full depth-based node budget
\(M=2^D-1\). Each configuration has a 600-second time limit.
CART is included as a fast greedy baseline. \texttt{Adaptive STreeD}
uses a single total time budget covering feature proposal, projection,
aggregation, and all inner \texttt{Weighted STreeD} solves.
To focus the computational study on scalability, all optimization-based
methods use training accuracy as the common objective. This experimental
choice does not restrict the weighted reformulation to accuracy.
Our implementation also includes weighted variants for $F_1$-score
optimization and cost-sensitive classification.

\textbf{Computing environment and status.}
The CART and Random Forest (RF) algorithms were implemented in
Python 3.12 using scikit-learn \citep{pedregosa2011}. All experiments
were conducted on a 64-bit Windows system with an Intel Core
i9-14900KF at 3.20 GHz and 64 GB of memory. 
The implementations of \texttt{Adaptive STreeD} and \texttt{Weighted STreeD}, together with reproducible examples, are publicly available at
\url{https://github.com/Tommytutu/AdaptiveSTreeD}.
For full-feature STreeD and DL8.5, \emph{Optimal} denotes a
certified optimum over the full candidate feature set. For fixed-candidate
Weighted STreeD, it denotes a certified optimum for the specified
candidate set. Adaptive STreeD does not in general provide a
full-feature optimality certificate; for this method, we report whether
a feasible incumbent is returned and whether the total time limit is reached.
A run with no feasible return before the time or memory limit is reported
as \emph{Infeasible}; this is an experimental status and does not imply
mathematical infeasibility.

\subsection{Overall Computational Performance}
\label{subsec:overall-performance}
This subsection addresses \textbf{Q1} by comparing computational
scalability and predictive performance across CART,
\texttt{Adaptive STreeD}, full-feature \texttt{STreeD}, and DL8.5
at depths 2--7 under a common 600-second time limit.
Full-feature \texttt{STreeD} serves as the direct reference because
\texttt{Adaptive STreeD} is built on the same dynamic-programming
framework, while DL8.5 provides an additional dynamic-programming
baseline for finite-budget optimal-tree search
\citep{aglin2020dl85}.

\textbf{Scalability.}
Results are matched by data set, random split, and maximum depth.
CART and \texttt{Adaptive STreeD} are evaluated at all depths.
Full-feature \texttt{STreeD} returns feasible trees only through
depth 4, and its certification rate drops sharply at depth 4.
At depths 5--7, runs that reach the time or memory limit without a
feasible tree are reported as \emph{Infeasible}. DL8.5 returns only
uncertified incumbents in all 75 runs at these depths. Comparisons at
larger depths are therefore interpreted under the common computational
budget rather than against consistently certified full-feature optima.

\begin{figure}[t]
\centering
\includegraphics[width=0.9\textwidth]
{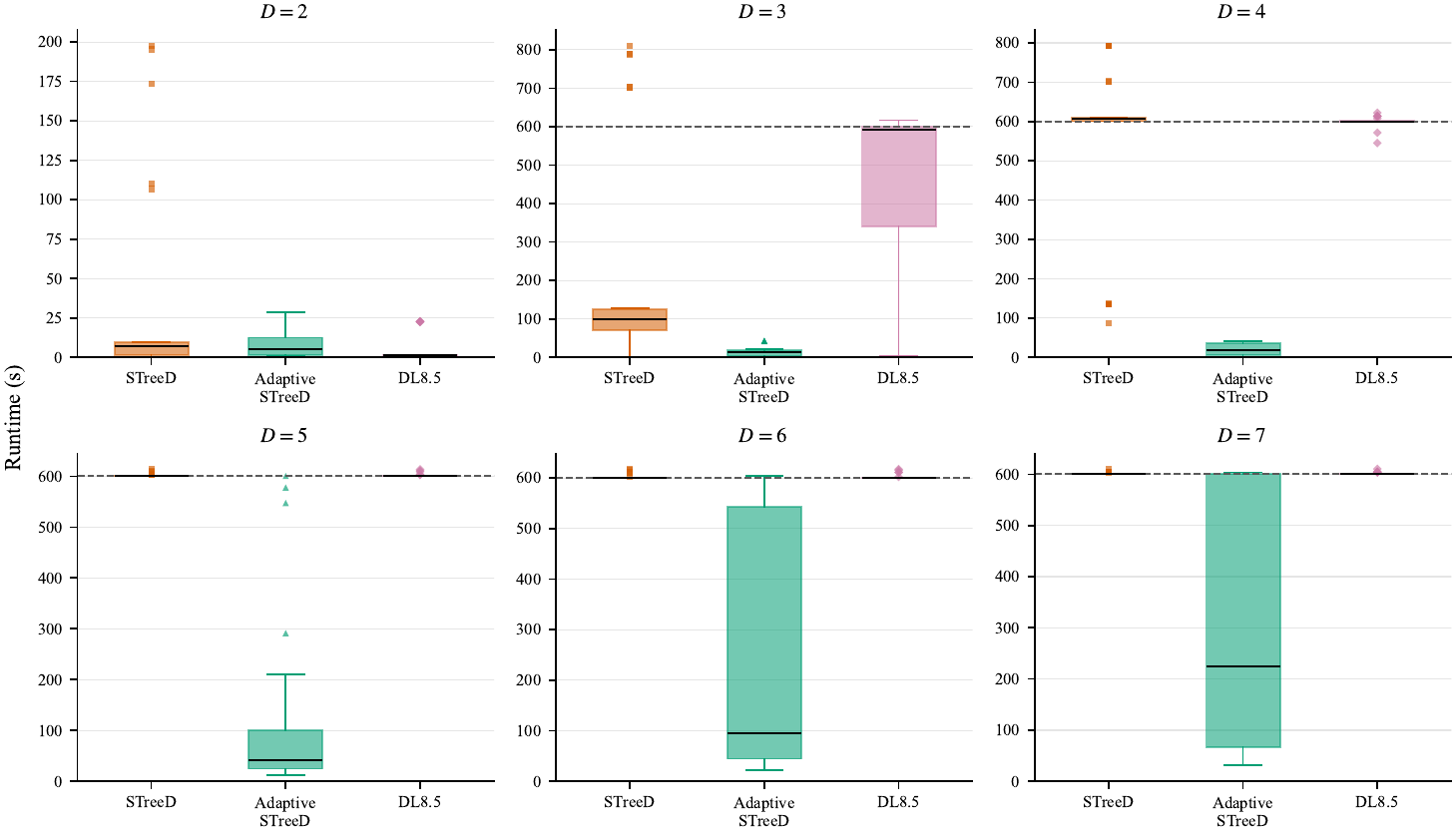}
\caption{Runtime distributions by maximum tree depth. Boxes show
interquartile ranges and center lines show medians. The dashed line
marks the 600-second time limit.}
\label{fig:overall-runtime}
\end{figure}

Figure~\ref{fig:overall-runtime} reports median runtimes and
interquartile ranges. At depths 2--4, median runtimes for
\texttt{Adaptive STreeD} are 5.38, 12.70, and 17.84 seconds, compared
with 7.19, 98.48, and 607.11 seconds for full-feature
\texttt{STreeD}. The corresponding runtime ratios are
\(1.34\times\), \(7.75\times\), and \(34.02\times\).
The depth-4 ratio is capped because many full-feature runs reach the
time or memory limit. The widening gap with depth reflects the growing
cost of full-feature search, whereas at depth 2 the overhead of feature
proposal and aggregation offsets much of the benefit.

At depths 5--7, \texttt{Adaptive STreeD} returns a feasible tree in
all 25 runs at each depth, with median runtimes of 42.08, 94.91, and
223.94 seconds. Among these runs, 1, 6, and 9 reach the time limit,
respectively. Full-feature \texttt{STreeD}, in contrast, fails to
return a feasible tree before the time or memory limit at these
depths. Candidate restriction therefore extends the range of depths
for which a feasible tree can be obtained under the stated
computational protocol, although the computational burden still grows
with depth.

DL8.5 is faster than \texttt{Adaptive STreeD} at depth 2, reflecting
the overhead of the adaptive procedure on easier instances. At depths
3 and 4, 11 of 25 and 23 of 25 DL8.5 runs reach the time limit, and
all 75 runs at depths 5--7 are time-limited. These solutions are
therefore treated as uncertified incumbents. Overall, the relative
computational advantage of \texttt{Adaptive STreeD} becomes more
pronounced as depth and problem difficulty increase.

\begin{figure}[t]
\centering

\begin{subfigure}[t]{0.9\textwidth}
\centering
\includegraphics[width=\linewidth]
{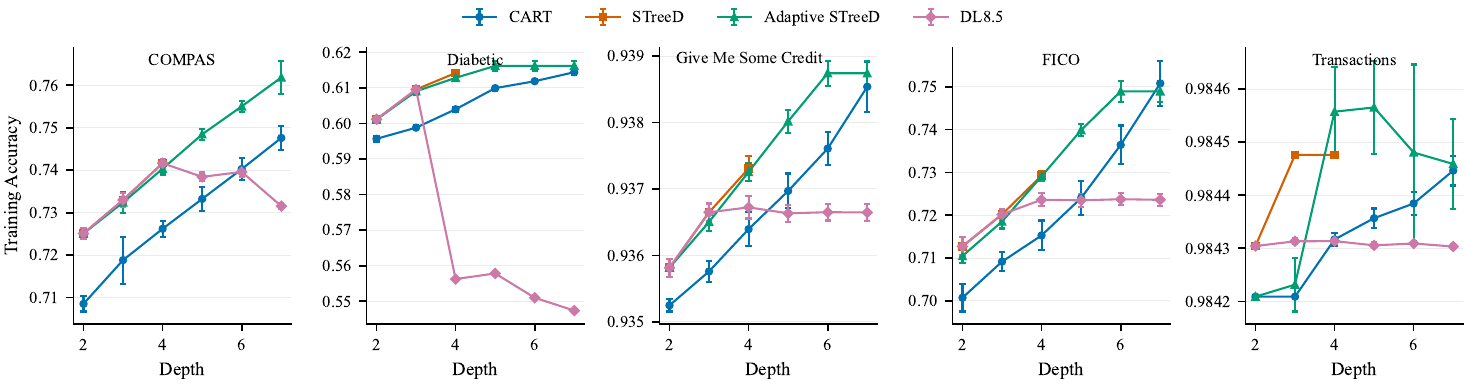}
\end{subfigure}

\vspace{0.05em}

\begin{subfigure}[t]{0.9\textwidth}
\centering
\includegraphics[width=\linewidth]
{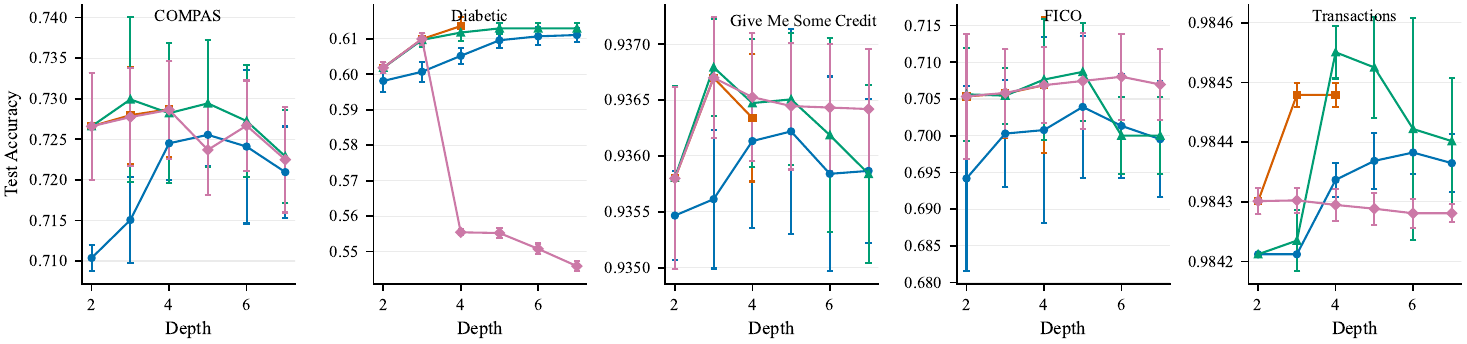}
\end{subfigure}

\caption{Training and test accuracy by data set and maximum tree
depth. Points show five-split means and error bars show sample
standard deviations.}
\label{fig:overall-predictive}
\end{figure}

\textbf{Predictive performance.}
Figure~\ref{fig:overall-predictive} reports five-split means with
sample-standard-deviation error bars. At depths 2--4,
\texttt{Adaptive STreeD} closely matches the available full-feature
\texttt{STreeD} results. Across the 75 matched
data-set--split--depth configurations, the training-accuracy
difference is at most 0.1 percentage points in 60 cases and at most
0.2 percentage points in 67 cases. The average
Adaptive-minus-\texttt{STreeD} difference is \(-0.0553\) percentage
points for training accuracy and \(0.08\) percentage points for
test accuracy. These differences are descriptive, and no statistical
significance is claimed.

For COMPAS, Diabetic, Give Me Some Credit, and FICO,
\texttt{Adaptive STreeD} generally achieves predictive performance
comparable to the other optimal-tree methods over the depths for which
matched results are available. The small test-accuracy differences
indicate that the scalability improvement does not come with a
systematic loss in predictive performance.

A different pattern appears for Transactions, the largest data set
with 786,363 observations. At depth 4, \texttt{Adaptive STreeD}
achieves better predictive performance than full-feature
\texttt{STreeD} and DL8.5. The full-feature search is particularly
difficult on this large instance, and many competing runs are limited
by the 600-second budget. Under the same budget,
\texttt{Adaptive STreeD} searches a reduced feature space and obtains
a tree with higher predictive performance. This is a finite-budget
advantage and does not imply that a certified full-feature optimum
would have lower predictive performance.

At greater depths, the pooled mean training accuracy of
\texttt{Adaptive STreeD} increases from \(80.08\%\) at depth 4 to
\(80.54\%\), \(80.87\%\), and \(81.00\%\) at depths 5--7.
Full-feature \texttt{STreeD} provides no comparable feasible solutions
at these depths, while DL8.5 returns time-limited incumbents. Test
accuracy, however, is not monotone in depth: the best five-split mean
occurs at depth 3 for COMPAS and Give Me Some Credit, depth 4 for
Diabetic and Transactions, and depth 5 for FICO. This pattern is
consistent with overfitting, as deeper trees can improve the training
objective without necessarily improving out-of-sample performance.

For \textbf{Q1}, the matched results indicate that
\texttt{Adaptive STreeD} scales better as tree depth and problem
difficulty increase. Its predictive performance remains comparable
to the evaluated optimal-tree methods under the same computational
budget. On the largest data set, the reduced search also yields higher
predictive performance within that budget.

\subsection{Fixed-Candidate Weighted Aggregation}
\label{subsec:weighted-ablation}

This subsection addresses \textbf{Q2} by isolating the computational
effect of weighted unique-data aggregation. A 500-tree random forest
ranks the binary features, and matched \texttt{STreeD} and
\texttt{Weighted STreeD} runs receive the same top
\(K\in\{10,20,30,40,50\}\) features, training split, depth,
objective, and 600-second time limit. Thus, the two methods solve the
same fixed-candidate tree problem and differ only in the
representation of the training records. Standard \texttt{STreeD}
retains all training records, whereas \texttt{Weighted STreeD} merges
duplicate projected feature--label records into weighted
representatives.

For each matched pair, we report
\begin{equation}
\label{eq:weighted-ablation-measures}
\mathrm{Compression}
=
\frac{n}{u_S},
\qquad
\mathrm{Speedup}
=
\frac{T_{\mathrm{STreeD}}}
     {T_{\mathrm{W\text{-}STreeD}}}.
\end{equation}
A speedup above one favors \texttt{Weighted STreeD}. Ratios involving
time-limited runs are interpreted as capped comparisons.

\begin{table}[t]
\centering
\caption{Weighted aggregation ablation on five binarized data sets}
\label{tab:weighted-aggregation-five-data-summary}
\begin{threeparttable}
\footnotesize
\setlength{\tabcolsep}{3pt}
\begin{tabular}{@{}lrrrrrrr@{}}
\toprule
Data set
& Pairs
& Mean comp.
& Mean sp.
& Max sp.
& Faster
& Same
& Better \\
\midrule
COMPAS
& 30 & 10.19 & 2.44 & 7.48
& 23/30 & 30/30 & 0 \\
Diabetic
& 30 & 39.05 & 3.83 & 36.68
& 18/30 & 28/30 & 2 \\
Give Me Some Credit
& 30 & 281.68 & 9.53 & 40.02
& 20/30 & 27/30 & 3 \\
FICO
& 30 & 98.78 & 2.14 & 7.96
& 19/30 & 30/30 & 0 \\
Transactions
& 30 & 538.46 & 23.05 & 121.41
& 20/30 & 23/30 & 7 \\
\midrule
\textbf{All}
& 150 & 193.63 & 8.20 & 121.41
& 100/150 & 138/150 & 12 \\
\bottomrule
\end{tabular}
\begin{tablenotes}[flushleft]
\footnotesize
\item \textit{Note.}
``Mean comp.'' and ``Mean sp.'' denote arithmetic-mean compression
and runtime speedup, respectively. ``Faster'' denotes a recorded
speedup greater than one. ``Same'' indicates equal training
misclassification, while ``Better'' indicates lower training
misclassification for \texttt{Weighted STreeD}.
\end{tablenotes}
\end{threeparttable}
\end{table}

\begin{figure}[t]
\centering
\begin{subfigure}[t]{0.49\textwidth}
\centering
\includegraphics[width=\linewidth]
{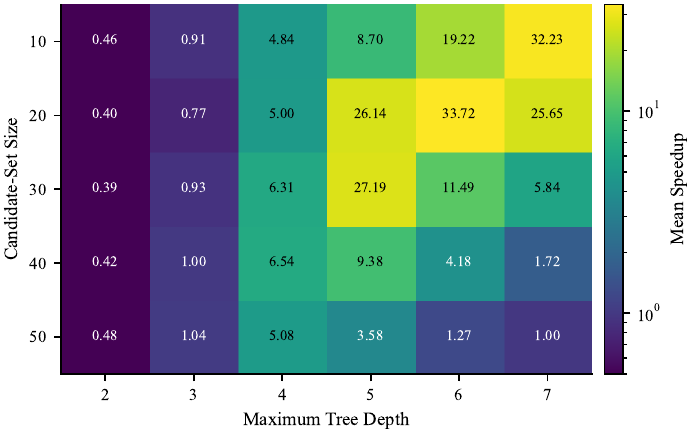}
\caption{Mean speedup by \(K\) and depth.}
\end{subfigure}
\hfill
\begin{subfigure}[t]{0.49\textwidth}
\centering
\includegraphics[width=\linewidth]
{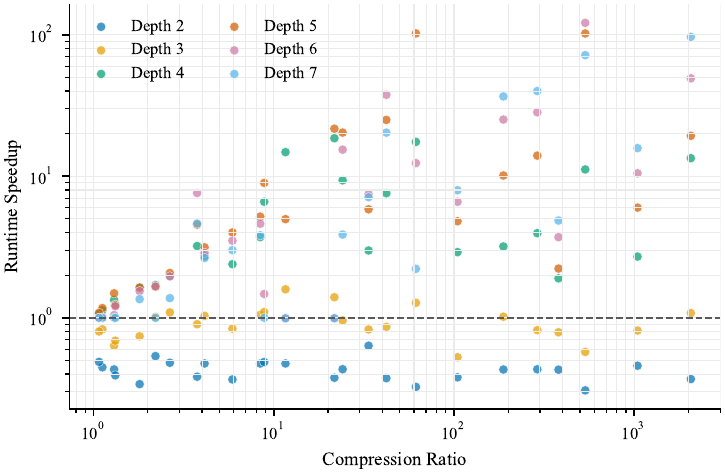}
\caption{Compression ratio and matched runtime speedup.}
\end{subfigure}
\caption{Effect of weighted unique-data aggregation under fixed
candidate sets. The dashed line in panel (b) indicates equal recorded
runtime between standard STreeD and Weighted STreeD.}
\label{fig:weighted-ablation}
\end{figure}

\textbf{Computational effect.}
Table~\ref{tab:weighted-aggregation-five-data-summary} and
Figure~\ref{fig:weighted-ablation} show that
\texttt{Weighted STreeD} has lower recorded runtime in 100 of 150
matched cases. All 50 cases in which it is not faster occur at
depths 2 and 3, where the underlying optimization problems are
relatively easy and aggregation overhead is more noticeable. Across
all cases, the arithmetic-mean speedup is \(8.20\times\), with a
maximum of \(121.41\times\).

Compression and speedup vary across data sets. Transactions has the
largest mean compression ratio, 538.46, and the largest mean speedup,
\(23.05\times\), while Give Me Some Credit achieves a mean speedup of
\(9.53\times\). In contrast, COMPAS has a smaller mean compression
ratio of 10.19 and a more moderate mean speedup of \(2.44\times\).

Depth and candidate-set size further explain the runtime differences.
At depths 2 and 3, the additional cost of aggregation, initialization,
and weight management can offset the savings from shorter record
scans. From depth 4 onward, \texttt{Weighted STreeD} is faster in all
matched cases, as repeated record processing across more
dynamic-programming states becomes more important. The largest gains
also tend to occur for smaller \(K\), because projection onto fewer
features creates more duplicate records. As \(K\) increases,
\(u_S\) generally approaches \(n\), reducing the benefit of
aggregation.

The compression ratio and runtime improvement are positively related,
but not proportional. Compression and log speedup have a Spearman
correlation of 0.342. This is consistent with the complexity analysis:
\(n/u_S\) measures the reduction in sample-dependent work, whereas
total runtime also includes feature enumeration, caching, pruning,
and solver overhead. Therefore, a large compression ratio does not
imply an equally large wall-clock speedup.

\textbf{Solution quality.}
The two methods solve the same fixed-candidate optimization problem,
so weighted aggregation does not change its optimum. Training
misclassification is identical in 138 of 150 matched cases. All 12
differences occur in time-limited runs, where
\texttt{Weighted STreeD} obtains a better incumbent before
termination. When both methods certify optimality,
Proposition~\ref{prop:objective-equivalence-revised} requires their
optimal objective values to agree.

For \textbf{Q2}, aggregation is most effective when projection creates
many duplicate records and tree depth makes repeated scanning costly.

\subsection{Sensitivity to Candidate-Set Capacity}
\label{subsec:sen_can_fea}

This subsection addresses \textbf{Q3} by examining how the
candidate-set capacity \(K\) affects computational cost and
predictive performance. We evaluate
\(K\in\{10,15,20,25,30,35,40\}\) at depths 2--7 on five data sets
and five train--test splits. Runtime ratios and test-accuracy
differences are computed within matched blocks using \(K=20\) as
the reference. Of the 1,050 planned runs, 1,043 return results;
missing runs are not imputed. Complete numerical results are reported
in Appendix~\ref{app:candidate-capacity-results}.

Seven configurations are unavailable. These missing runs occur only
at larger candidate capacities and greater depths and are therefore
not treated as random. Accuracy differences are computed only for
matched blocks in which both the target capacity and $K=20$ return a
feasible tree.

\begin{figure}[t]
\centering
\begin{subfigure}[t]{0.49\textwidth}
\centering
\includegraphics[width=\linewidth]
{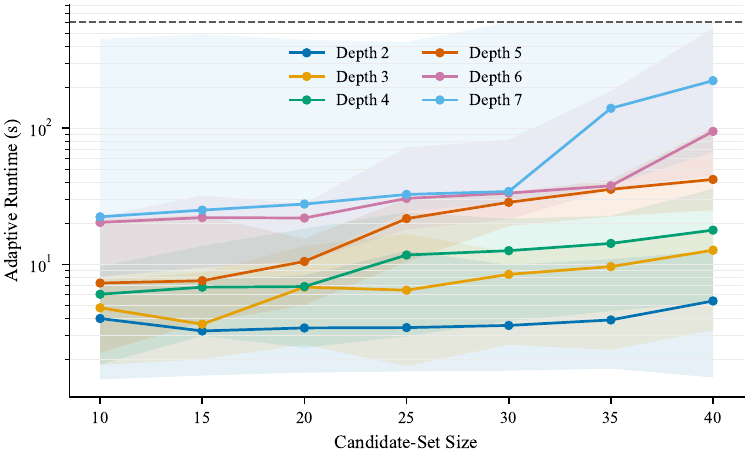}
\caption{Runtime sensitivity.}
\end{subfigure}
\hfill
\begin{subfigure}[t]{0.49\textwidth}
\centering
\includegraphics[width=\linewidth]
{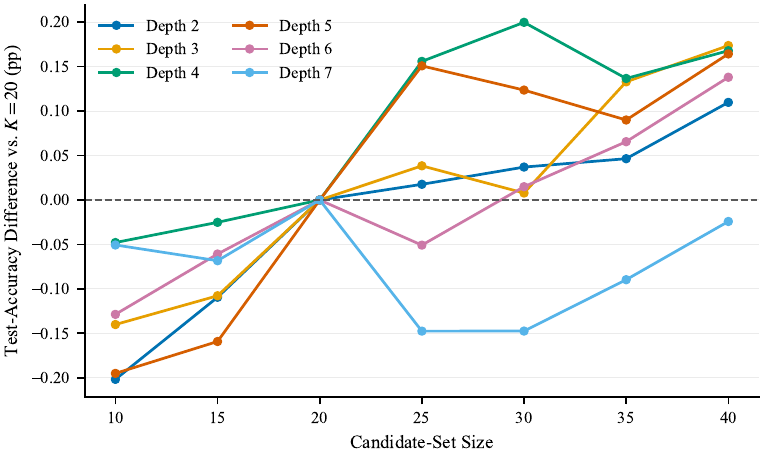}
\caption{Matched test-accuracy difference.}
\end{subfigure}
\caption{Sensitivity to candidate-set capacity \(K\). Runtime curves
show depth-specific medians on a logarithmic scale, with shaded
interquartile ranges. Accuracy curves report mean matched
test-accuracy differences relative to \(K=20\) in percentage points.}
\label{fig:candidate-sensitivity}
\end{figure}

\textbf{Computational cost.}
Figure~\ref{fig:candidate-sensitivity}(a) shows that runtime generally
increases with \(K\), and the effect becomes stronger as tree depth
increases. At \(K=20\), all 150 runs return results and the pooled
median runtime is 10.48 seconds. At \(K=40\), 144 runs return
results, the pooled median runtime increases to 32.03 seconds, and
the median matched runtime ratio relative to \(K=20\) is
\(2.433\times\).

Across the full capacity sequence, pooled median runtime increases
from 7.95 seconds at \(K=10\) to 20.89 seconds at \(K=30\) and
32.03 seconds at \(K=40\). All 150 runs return results through
\(K=30\), compared with 149 at \(K=35\) and 144 at \(K=40\).
The runtime increase is especially pronounced at greater depths. For
example, at depth 6, the median matched runtime ratio for \(K=40\)
relative to \(K=20\) reaches \(7.75\times\).

This behavior is consistent with the complexity analysis. A larger
candidate set increases the number of split features considered at
each dynamic-programming state. It can also create more distinct
projected records, reducing the amount of compression available to
\texttt{Weighted STreeD}. Both effects become more important as tree
depth increases.

\textbf{Predictive performance.}
Figure~\ref{fig:candidate-sensitivity}(b) shows that predictive
performance changes much less than runtime. Pooled mean training
accuracy is 0.7942, 0.7965, and 0.8029 for
\(K=10\), \(20\), and \(40\), respectively, while pooled mean test
accuracy is 0.7892, 0.7905, and 0.7946. Relative to \(K=20\), the
mean matched test-accuracy difference ranges from \(-0.127\)
percentage points at \(K=10\) to \(+0.122\) percentage points at
\(K=40\).

Increasing \(K\) therefore provides modest average improvements in
predictive performance compared with its effect on runtime. The
depth-specific differences are also not monotone. A larger candidate
set gives the algorithm access to more features, but the additional
coverage does not necessarily improve test accuracy at every depth.

For \textbf{Q3}, the results identify a tradeoff between feature
coverage and computational cost. Increasing \(K\) provides more
candidate features and modestly improves average predictive
performance, but at a much larger computational cost. The completion
rate also decreases at the largest capacities, especially for deeper
trees. In these experiments, \(K=20\) provides a computationally
efficient operating point, whereas \(K=40\) provides broader feature
coverage and slightly higher solution quality at greater
computational cost.

\section{Conclusion}
\label{sec:conclusion}

This paper develops two complementary mechanisms for improving the
scalability of dynamic-programming-based optimal classification trees.
\texttt{Weighted STreeD} merges duplicate projected records into
weighted representatives while preserving the fixed-candidate
optimization problem under the stated assumptions.
\texttt{Adaptive STreeD} further combines bounded candidate sets,
feature refinement, and \texttt{Weighted STreeD}. Each certified
inner solution is optimal for its current candidate set, while the
outer feature search remains heuristic over the full feature space.

Across five data sets, \texttt{Adaptive STreeD} requires less runtime
than full-feature \texttt{STreeD} in matched comparisons at depths
2--4 and continues to return feasible trees at greater depths. Its
predictive performance remains comparable to the evaluated optimal-tree
methods under the same computational budget. On the largest data set,
Transactions, it also achieves higher predictive performance in the
finite-budget comparison with the full-feature methods. The
fixed-candidate ablation shows that \texttt{Weighted STreeD} achieves
an average speedup of \(8.20\times\) and a maximum of
\(121.41\times\), with larger gains for deeper trees and more
compressible data. The sensitivity analysis further identifies a
tradeoff in the candidate capacity \(K\): increasing \(K\) improves
feature coverage and slightly improves solution quality, but also
increases runtime, especially for deeper trees.

The main limitation is that \texttt{Adaptive STreeD} does not
guarantee full-feature optimality because the feature-proposal
procedure may omit features used by an optimal tree. Its performance
also depends on the candidate capacity, time allocation, and stopping
rules, and reducing the feature and sample spaces does not remove the
combinatorial dependence on tree depth. In addition, the current
analysis gives explicit recursions and complexity bounds
only for binarized candidate features under the training-accuracy objective.
Proposition~\ref{prop:objective-equivalence-revised} separately
establishes fixed-candidate equivalence for separable tasks that satisfy
the information-preservation condition.

Future work will investigate safe feature-screening rules, stronger
full-space bounds, adaptive candidate capacities, and more efficient
memory and parallelization strategies. Extensions to continuous
features, task-specific recursions and complexity bounds
for objectives beyond training accuracy, robust formulations, and broader
high-stakes applications are also left for future work.

\begingroup
\setlength{\bibsep}{0pt}
\bibliographystyle{plainnat}
\bibliography{references_revised}
\endgroup

\newpage
\appendix
\renewcommand{\theHsection}{appendix.\Alph{section}}

\section{Binning Sensitivity}
\label{app:binning-sensitivity-weighted-unique}

The main experiments discretize each continuous feature using at most
100 quantile bins before cumulative binary encoding. The value 100 is
an upper bound rather than a requirement that every feature produce
100 distinct bins. Tied values and repeated quantiles may produce
fewer intervals and, consequently, fewer binary thresholds. This
resolution defines a common candidate threshold universe for all
methods while limiting the threshold information removed before
optimization. The 100-bin setting is not assumed to minimize runtime
or maximize predictive accuracy. It also serves a different purpose
from the candidate capacity \(K\): the binning cap determines the
feature universe during preprocessing, whereas \(K\) limits the active
feature set during Adaptive STreeD.

We examine the effect of preprocessing resolution by applying
Weighted STreeD to representations constructed with
\(B\in\{5,10,20,50,100\}\) maximum quantile bins. Adaptive candidate
refinement is not used in this experiment, so the comparison isolates
the effect of \(B\). We evaluate five data sets at depths
\(D\in\{2,3,4\}\), giving 15 configurations for each value of \(B\)
and 75 configurations in total. All runs use the same fixed 80/20
train--test split with random state 42. Using a fixed split avoids
variation from repeated data partitioning, but the resulting accuracy
differences should be interpreted as descriptive rather than as
estimates across random splits.

\begin{figure}[t]
\centering

\begin{subfigure}[t]{0.49\textwidth}
\centering
\includegraphics[width=\linewidth]
{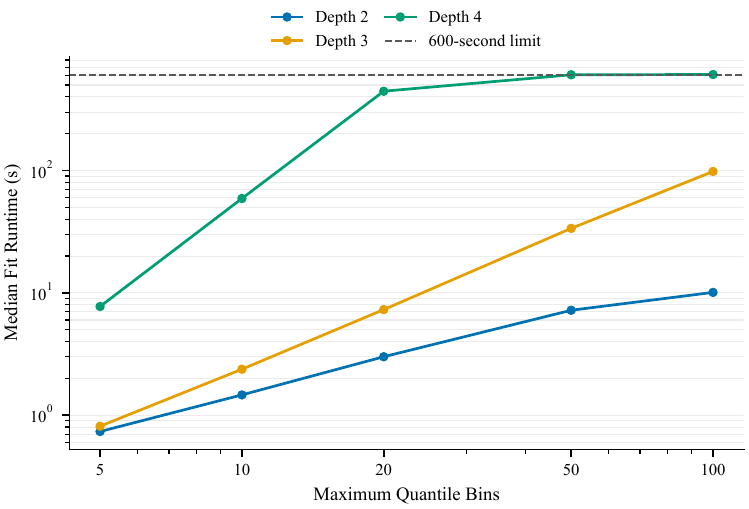}
\caption{Median recorded runtime by depth.}
\end{subfigure}
\hfill
\begin{subfigure}[t]{0.49\textwidth}
\centering
\includegraphics[width=\linewidth]
{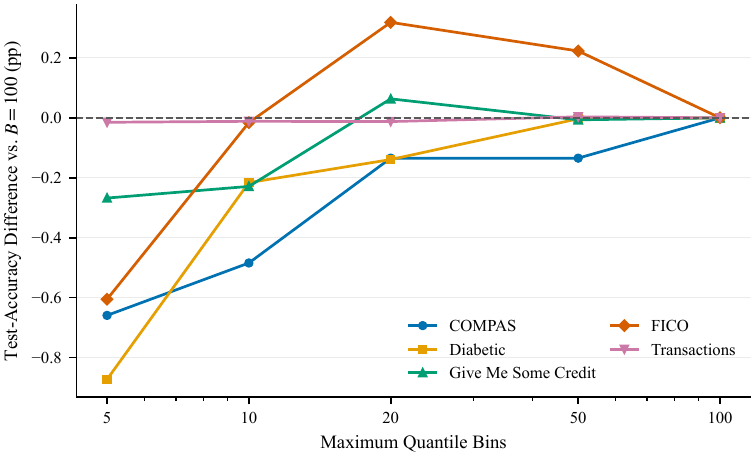}
\caption{Test-accuracy difference relative to \(B=100\).}
\end{subfigure}

\caption{Sensitivity to the maximum number of quantile bins.
Panel~(a) reports the median recorded runtime over the five data sets
at each depth; the dashed line marks the nominal 600-second limit.
Panel~(b) reports the mean test-accuracy difference relative to
\(B=100\) for each data set, averaged over depths 2--4. Positive
values favor the indicated binning level. Time-limited runs contribute
their best feasible incumbents.}
\label{fig:appendix-binning-sensitivity}
\end{figure}

Each configuration maximizes training accuracy, uses the complete
depth-based node budget, and has a nominal runtime limit of 600
seconds. A time-limited run contributes its best feasible incumbent
when one is available. Table~\ref{tab:direct-binning-weighted-unique}
reports the pooled results, while
Figure~\ref{fig:appendix-binning-sensitivity} separates the runtime
effect by depth and the test-accuracy effect by data set.

\begin{table}[H]
\centering
\caption{Sensitivity of Weighted STreeD to the maximum number of quantile bins}
\label{tab:direct-binning-weighted-unique}
\begin{threeparttable}
\footnotesize
\setlength{\tabcolsep}{2.5pt}
\renewcommand{\arraystretch}{1.08}
\begin{tabular}{@{}lrrrrrrr@{}}
\toprule
Max bins & Mean binary feat. & Median runtime (s) & $D=4$ runtime (s) & Limit share & Mean train acc. & Mean test acc. & Median comp. \\
\midrule
5 & 100.8 & 1.50 & 7.74 & 6.7\% & 0.79291 & 0.78772 & 1.28 \\
10 & 139.0 & 2.96 & 59.07 & 13.3\% & 0.79526 & 0.79065 & 1.07 \\
20 & 201.6 & 7.30 & 444.87 & 13.3\% & 0.79616 & 0.79275 & 1.06 \\
50 & 344.2 & 33.67 & 606.12 & 26.7\% & 0.79663 & 0.79273 & 1.02 \\
100 & 512.0 & 119.71 & 609.26 & 33.3\% & 0.79670 & 0.79256 & 1.01 \\
\bottomrule
\end{tabular}
\begin{tablenotes}[flushleft]
\footnotesize
\item \textit{Note.} Each row aggregates 15 successful runs (five data sets and depths $D\in\{2,3,4\}$) using \texttt{Weighted STreeD}. Binary-feature counts and accuracies are means; runtimes and compression factors are medians. ``Limit share'' is the fraction of runs with fit runtime at least 599 seconds. The $D=4$ column is the median over the five depth-4 runs. The bin count is a cap; repeated quantiles can yield fewer distinct thresholds.
\end{tablenotes}
\end{threeparttable}
\end{table}

Increasing the binning resolution substantially raises computational
cost. As \(B\) increases from 5 to 100, the average number of binary
features increases from 100.8 to 512.0, corresponding to a
\(5.1\times\) increase. The pooled median recorded runtime rises from
1.50 to 119.71 seconds, while the depth-4 median increases from 7.74
to 609.26 seconds. The fraction of runs with a recorded runtime of at
least 599 seconds increases from \(6.7\%\) to \(33.3\%\).

Recorded runtimes may slightly exceed the nominal 600-second limit
because termination checks and result collection introduce a small
amount of additional time. The largest depth-4 observations should
therefore be interpreted as censored by the computational budget
rather than as uncensored runtime measurements. Nevertheless, the
measurements indicate that finer binning increases computational cost.

Finer binning affects both sources of dynamic-programming cost. It
creates more candidate split features and reduces the number of
duplicate projected records. The median compression ratio consequently
decreases from 1.28 at \(B=5\) to 1.01 at \(B=100\). Increasing the
preprocessing resolution therefore expands the feature search space
and leaves fewer duplicate records for Weighted STreeD to aggregate.

The predictive effect is much smaller and is not monotone. Mean test
accuracy increases by approximately 0.50 percentage points from
\(B=5\) to \(B=20\). The pooled mean test accuracies at \(B=20\) and
\(B=100\) are 0.79275 and 0.79256, respectively, corresponding to a
difference of only 0.02 percentage points in favor of \(B=20\).
Hence, the sensitivity study does not identify 100 bins as a
predictive optimum.

Instead, the results support the intended role of Bin100 as a
conservative preprocessing cap. It preserves a relatively rich set of
candidate thresholds and applies the same encoded representation to
all methods. Adaptive candidate refinement is then responsible for
controlling the feature dimension considered during optimization. A
smaller cap, such as \(B=20\), provides a computationally attractive
alternative and achieves similar predictive performance in this
fixed-split experiment, but it defines a coarser candidate universe
and may discard thresholds before optimization. Because the analysis
uses only one train--test split, broader conclusions about the
predictive effect of \(B\) would require evaluation across repeated
random splits.

\section{Supplementary Method Details}
\label{app:method-details}

\subsection{Notation}
\label{app:notation}
\begin{table}[H]
\centering
\caption{Notation used in the weighted unique-data dynamic program.}
\label{tab:weighted-dp-notation}

\begin{threeparttable}
\footnotesize
\renewcommand{\arraystretch}{1.08}

\begin{tabular}{
    p{1.6cm}
    p{7.0cm}
    p{4.1cm}}
\toprule
Symbol & Meaning & Scope \\
\midrule

$\mathcal{I}$
& Original training set
& Fixed input data \\

$\mathcal{F}$
& Full feature index set $\{1,\ldots,p\}$
& Full post-binarization space \\

$S$
& Candidate split-feature set
& Fixed within one DP solve \\

$q$
& Candidate-set size, $q=|S|$
& Fixed within one DP solve \\

$D$
& Maximum tree depth
& Feasible tree class \\

$M$
& Maximum number of internal split nodes
& Feasible tree class \\

$d$
& Remaining depth
& DP state variable \\

$m$
& Remaining internal-node budget
& DP state variable \\

$A$
& Set of original records reaching a state
& Baseline STreeD \\

$B$
& Set of weighted records reaching a state
& Weighted STreeD \\

$n_A$
& Number of original records in state $A$
& Baseline state \\

$u_B$
& Number of weighted records in state $B$
& Weighted state \\

$\eta_i$
& Record-level information required by the task
& Aggregation key \\

$\boldsymbol{\phi}_i(T)$
& Record $i$'s contribution to the task statistics under tree $T$
& Separable-task equivalence \\

$g_i(S)$
& Aggregation key $(\mathbf{x}_{iS},y_i,\eta_i)$
& Fixed candidate set \\

$\mathcal{U}(S)$
& Weighted unique data induced by $S$
& Fixed candidate set \\

$u_S$
& Number of weighted unique records
& Fixed candidate set \\

$H_D(q)$
& Conservative path-state bound
& Complexity analysis \\

\bottomrule
\end{tabular}

\begin{tablenotes}[flushleft]
\footnotesize
\item \textit{Note.}
The candidate set and post-binarized feature representation remain
fixed within each dynamic-programming solve.
\end{tablenotes}

\end{threeparttable}
\end{table}

\subsection{Baseline STreeD Recursion}
\label{app:baseline-recursion}
\begin{algorithm}[H]
\caption{Simplified STreeD Recursion for Fixed Candidate Features}
\label{alg:streed-baseline}
\footnotesize
\begin{algorithmic}[1]
\Require Record set $A$, candidate features $S$, remaining depth $d$,
remaining internal-node budget $m$, cache $\mathcal{C}$

\If{$(A,d,m)\in\mathcal{C}$}
    \State \Return $\mathcal{C}(A,d,m)$
\EndIf

\State $C^\star\gets\mathrm{Leaf}(A)$
\State $T^\star\gets$ best leaf for $A$

\If{$d=0$ or $m=0$}
    \State $\mathcal{C}(A,d,m)\gets(C^\star,T^\star)$
    \State \Return $(C^\star,T^\star)$
\EndIf

\For{$f\in S$}
    \State
    $A_0(f)\gets\{i\in A:x_{if}=0\}$
    \State
    $A_1(f)\gets\{i\in A:x_{if}=1\}$

    \If{$A_0(f)=\emptyset$ or $A_1(f)=\emptyset$}
        \State \textbf{continue}
    \EndIf

    \For{$m_0=0,\ldots,m-1$}
        \State $m_1\gets m-1-m_0$

        \State
        $(C_0,T_0)
        \gets
        \mathrm{STreeD}
        (A_0(f),S,d-1,m_0,\mathcal{C})$

        \State
        $(C_1,T_1)
        \gets
        \mathrm{STreeD}
        (A_1(f),S,d-1,m_1,\mathcal{C})$

        \If{$C_0+C_1<C^\star$}
            \State $C^\star\gets C_0+C_1$
            \State $T^\star\gets(f,T_0,T_1)$
        \EndIf
    \EndFor
\EndFor

\State $\mathcal{C}(A,d,m)\gets(C^\star,T^\star)$
\State \Return $(C^\star,T^\star)$
\end{algorithmic}
\end{algorithm}

\section{Supplementary Experimental Results}
\label{app:supplementary-experiments}

\subsection{Complete Candidate-Capacity Results}
\label{app:candidate-capacity-results}
\begin{table}[H]
\centering
\caption{Sensitivity of Adaptive STreeD to candidate-set capacity}
\label{tab:candidate-k-sensitivity}
\begin{threeparttable}
\footnotesize
\begin{tabular}{rrrrrrr}
\toprule
$K$
& Successful
& Runtime
& Train acc.
& Test acc.
& Runtime/$K=20$
& Test diff. (pp) \\
\midrule
10
& 150
& 7.95 [2.29--22.30]
& $0.7942\pm0.1420$
& $0.7892\pm0.1449$
& 0.788
& -0.127 \\

15
& 150
& 9.29 [2.72--24.74]
& $0.7953\pm0.1415$
& $0.7896\pm0.1447$
& 0.897
& -0.088 \\

20
& 150
& 10.48 [3.35--27.44]
& $0.7965\pm0.1408$
& $0.7905\pm0.1443$
& 1.000
& 0.000 \\

25
& 150
& 17.99 [4.42--30.66]
& $0.7976\pm0.1408$
& $0.7907\pm0.1448$
& 1.110
& 0.027 \\

30
& 150
& 20.89 [6.39--33.48]
& $0.7982\pm0.1404$
& $0.7909\pm0.1447$
& 1.241
& 0.039 \\

35
& 149
& 27.45 [8.19--39.61]
& $0.7995\pm0.1403$
& $0.7916\pm0.1448$
& 1.563
& 0.064 \\

40
& 144
& 32.03 [11.49--82.55]
& $0.8029\pm0.1419$
& $0.7946\pm0.1465$
& 2.433
& 0.122 \\
\bottomrule
\end{tabular}
\begin{tablenotes}[flushleft]
\footnotesize
\item \textit{Note.}
Runtime is the median [interquartile range] in seconds over available
configurations. Accuracies are mean $\pm$ sample standard deviation.
Runtime ratios and test-accuracy differences are computed within
matched blocks relative to $K=20$. The reported runtime ratio is the
median of within-block ratios and is not the ratio of the pooled
median runtimes.
\end{tablenotes}
\end{threeparttable}
\end{table}

\section{Proofs}
\label{app:proofs}

\subsection{Proof of Proposition~\ref{prop:streed-time}}
\label{app:proof-streed-complexity}

Consider a cached state $(A,d,m)$, where $n_A=|A|$. Under the direct
state-scanning model, computing the class counts required by the best
leaf costs $O(|\mathcal{K}|n_A)$ in the stated conservative
accounting. Evaluating the $q$ binary candidate splits requires
$O(qn_A)$ additional work. For each candidate split, the remaining
internal-node budget can be divided between the two child subtrees in
at most $m$ ways, contributing $O(qm)$ work once the child values are
available. The total cost at the state is therefore
\[
O\!\left((|\mathcal{K}|+q)n_A+qm\right).
\]
Summing this quantity over all states in
$\mathcal{A}_{D,M}(S)$ gives
\eqref{eq:streed-time-state}.

To obtain the worst-case bound, consider a nonredundant path of
length $\ell$. Such a path fixes at most $\ell$ distinct binary
features and one outcome for each selected feature. The number of
possible path descriptions is therefore at most
\[
\binom{q}{\ell}2^\ell.
\]
Summing over $\ell=0,\ldots,\min\{D,q\}$ gives the path-state bound
$H_D(q)$ in \eqref{eq:path-state-bound}. Including the possible
remaining-depth and remaining-budget values gives at most
\[
(D+1)(M+1)H_D(q)
\]
cached states. Using $n_A\le n$ and $m\le M$ in the per-state cost
then gives \eqref{eq:streed-time-worst}.

For the space bound, a bitset representing a subset of the $n$
original records requires $O(n/b)$ machine words. The remaining
information stored with a state, including $d$, $m$, and its
dynamic-programming value, requires constant space. Summing over all
cached states gives
\eqref{eq:streed-space}. The input feature matrix, returned trees,
and auxiliary solver statistics are excluded, as stated in the
proposition.

\subsection{Proof of Proposition~\ref{prop:objective-equivalence-revised}}
\label{app:proof-objective-equivalence}

For each weighted record $u\in\mathcal{U}(S)$, let
\[
\mathcal{G}_u
=
\{i:g_i(S)=(\mathbf{z}_u,y_u,\eta_u)\}
\]
be the group of original observations represented by $u$. By
construction, $|\mathcal{G}_u|=w_u$. Every observation in
$\mathcal{G}_u$ has the same projected features, label, and record-level
task information. Moreover, any $T\in\mathcal{T}(S,D,M)$ uses only
features in $S$, so all observations in the group reach the same leaf
and receive the same prediction. Their sufficient-statistic
contributions are therefore identical; denote this common vector by
$\boldsymbol{\phi}_u(T)$. It follows that
\[
\sum_{i\in\mathcal{G}_u}\boldsymbol{\phi}_i(T)
=w_u\boldsymbol{\phi}_u(T).
\]
Because the groups $\mathcal{G}_u$ partition the original data, summing
over $u$ gives
\[
\sum_{i=1}^{n}\boldsymbol{\phi}_i(T)
=
\sum_{u\in\mathcal{U}(S)}w_u\boldsymbol{\phi}_u(T).
\]
Thus, every feasible tree has the same task statistics under the
original and weighted representations. The feasible tree class and all
tree-only terms are also unchanged. Applying the same, possibly
nonlinear, task objective and constraints therefore gives each tree the
same objective value and feasibility status in both representations.
The two fixed-candidate problems consequently have the same optimal
value and the same set of optimal trees.

\subsection{Proof of Corollary~\ref{prop:weighted-recursion-validity}}
\label{app:proof-weighted-recursion}

For a weighted state $B\subseteq\mathcal{U}(S)$, let
\[
A(B)
=
\bigcup_{u\in B}\mathcal{G}_u
\]
be the corresponding set of original observations. The weighted class
counts in $B$ equal the unweighted class counts in $A(B)$. Therefore,
the weighted leaf cost in \eqref{eq:weighted-best-leaf} equals the
unweighted leaf cost of the corresponding original state.

For every $f\in S$, the split of $B$ into $B_0(f)$ and $B_1(f)$
corresponds exactly to the split of $A(B)$ into its original
observation copies. In particular,
\[
A(B_j(f))
=
\left\{
i\in A(B):x_{if}=j
\right\},
\qquad j\in\{0,1\}.
\]

The result now follows by induction on the remaining depth and node
budget. When $d=0$ or $m=0$, both recursions return the same leaf
cost. For $d>0$ and $m>0$, assume that corresponding child states
have equal optimal values. The two recursions examine the same split
features and the same allocations satisfying
$m_0+m_1=m-1$. Their child-state values are equal by the induction
hypothesis, and their leaf alternatives are also equal. Taking the
minimum over the same alternatives therefore gives the same value at
the parent state.

Aggregation-preserving bounds and pruning rules do not change this
correspondence because they are computed from the same weighted
sufficient statistics. Hence, the weighted and unweighted recursions
return the same optimal misclassification count over
$\mathcal{T}(S,D,M)$. 

\subsection{Proof of Proposition~\ref{prop:weighted-streed-time}}
\label{app:proof-weighted-streed-complexity}

At a weighted state $(B,d,m)$, \texttt{Weighted STreeD} scans
$u_B=|B|$ weighted records. Computing weighted class counts and
evaluating the $q$ candidate splits require
\[
O\!\left((|\mathcal{K}|+q)u_B\right)
\]
work. Allocating the remaining internal-node budget between the child
subtrees contributes $O(qm)$. Summing the resulting per-state cost
over $\mathcal{B}_{D,M}(S)$ gives
\eqref{eq:weighted-streed-time-state}.

The logical path-state bound remains $H_D(q)$ because aggregation
changes the record representation but not the candidate features or
the possible feature--outcome paths. There are therefore at most
\[
(D+1)(M+1)H_D(q)
\]
depth- and budget-indexed states under the conservative accounting.
Using $u_B\le u_S$ and $m\le M$ gives
\eqref{eq:weighted-streed-time-worst}.

For the space bound, a bitset over the $u_S$ weighted records
requires $O(u_S/b)$ machine words, and the remaining information
stored with a state requires constant space. Summing over the cached
weighted states gives \eqref{eq:weighted-streed-space}. Storage for
the weighted input table, returned trees, and auxiliary solver
statistics is excluded. \hfill$\square$

\subsection{Proof of Proposition~\ref{prop:sample-dependent-reduction}}
\label{app:proof-sample-dependent-reduction}

Under the proposition's correspondence assumption, each logical
baseline state $(A,d,m)$ has a corresponding weighted state
$(B,d,m)$. The sample-scanning part of the baseline running time is
proportional to
\[
(|\mathcal{K}|+q)
\sum_{(A,d,m)\in\mathcal{A}_{D,M}(S)}
n_A,
\]
whereas the corresponding term for \texttt{Weighted STreeD} is proportional
to
\[
(|\mathcal{K}|+q)
\sum_{(B,d,m)\in\mathcal{B}_{D,M}(S)}
u_B.
\]
The common factor $|\mathcal{K}|+q$ cancels, giving exactly
\eqref{eq:time-reduction-exact}. The $qm$ budget-allocation terms are
not included because the proposition concerns only the
sample-scanning component.

Suppose that the compression ratio in each pair of corresponding
states is close to the root-level ratio:
\[
\frac{n_A}{u_B}
\approx
\frac{n}{u_S}.
\]
Equivalently, $n_A\approx(n/u_S)u_B$. Summing this relation over the
corresponding states gives
\[
\sum n_A
\approx
\frac{n}{u_S}\sum u_B.
\]
Substitution into \eqref{eq:time-reduction-exact} yields
\eqref{eq:complexity-reduction-summary}.

For bitset-based representations, every baseline state uses a bitset
over the universe of $n$ original records, whereas every weighted
state uses a bitset over the universe of $u_S$ weighted records.
Including constant state metadata gives the per-state ratio
\[
\frac{n/b+1}{u_S/b+1},
\]
which is \eqref{eq:space-reduction-exact}. When both $n/b$ and
$u_S/b$ are large, the additive constants are negligible and the
ratio approaches $n/u_S$.

\subsection{Proof of Proposition~\ref{prop:bounded-subproblem}}
\label{app:proof-bounded-subproblem}

At iteration $t$, apply
\eqref{eq:weighted-streed-time-worst} with candidate set $E_t$,
candidate-set size $q_t=|E_t|$, and weighted unique-data size
$u_t=|\mathcal{U}(E_t)|$. This directly gives
\eqref{eq:adaptive-inner-complexity}.

For ordinary classification with binarized candidate features, there
are at most $2^{q_t}$ projected feature vectors and
$|\mathcal{K}|$ possible labels. Hence,
\[
u_t
\le
\min\{n,|\mathcal{K}|2^{q_t}\}.
\]
Since $q_t\le K$,
\[
H_D(q_t)\le H_D(K),
\qquad
u_t\le
\min\{n,|\mathcal{K}|2^K\}.
\]
Moreover,
\[
|\mathcal{K}|+q_t
\le
|\mathcal{K}|+K,
\qquad
q_tM\le KM.
\]
Substituting these inequalities into
\eqref{eq:adaptive-inner-complexity} gives the uniform bound
\eqref{eq:bounded-core}.

This proposition begins after $E_t$ and
$\mathcal{U}(E_t)$ have been constructed. It therefore does not
include random-forest or CART fitting, projection, or aggregation.
Those terms are accounted for separately in
\eqref{eq:adaptive-total-cost}. 

\subsection{Proof of Proposition~\ref{prop:expected-approximation}}
\label{app:expected-approximation-proof}

Let
\[
\mathcal{B}_t
=
\bigcap_{s=1}^{t}\mathcal{A}_s^c
\]
be the event that the selected optimal support has not been covered
during the first $t$ completed iterations. Since
$\mathcal{B}_{t-1}$ is $\mathcal{H}_{t-1}$-measurable,
\begin{align*}
\Pr(\mathcal{B}_t)
&=
\mathbb{E}\!\left[
\mathbf{1}_{\mathcal{B}_{t-1}}
\mathbf{1}_{\mathcal{A}_t^c}
\right]
\\
&=
\mathbb{E}\!\left[
\mathbf{1}_{\mathcal{B}_{t-1}}
\Pr\!\left(
\mathcal{A}_t^c
\mid
\mathcal{H}_{t-1}
\right)
\right].
\end{align*}
On $\mathcal{B}_{t-1}$, condition
\eqref{eq:conditional-coverage} implies
\[
\Pr\!\left(
\mathcal{A}_t^c
\mid
\mathcal{H}_{t-1}
\right)
\le
1-\underline{\rho}_t.
\]
Therefore,
\[
\Pr(\mathcal{B}_t)
\le
(1-\underline{\rho}_t)
\Pr(\mathcal{B}_{t-1}).
\]
Iterating from $\Pr(\mathcal{B}_0)=1$ gives
\[
\Pr(\mathcal{B}_N)
\le
\prod_{t=1}^{N}
(1-\underline{\rho}_t).
\]

On $\mathcal{B}_N^c$, at least one active candidate set contains
$S_D^\star$. The corresponding reduced tree class contains
$T_D^\star$ and is a subset of the full-feature tree class. Its
optimal score is therefore exactly $Q_D^\star$. By assumption, that
reduced problem is solved to certified optimality, and incumbent
preservation retains the score $Q_D^\star$ through iteration $N$.

On $\mathcal{B}_N$, the baseline assumption gives
\[
\frac{Q_{A,N}}{Q_D^\star}\ge r_0.
\]
Consequently,
\begin{align*}
\mathbb{E}\!\left[
\frac{Q_{A,N}}{Q_D^\star}
\right]
&\ge
\Pr(\mathcal{B}_N^c)
+
r_0\Pr(\mathcal{B}_N)
\\
&=
1-(1-r_0)\Pr(\mathcal{B}_N)
\\
&\ge
1-
(1-r_0)
\prod_{t=1}^{N}
(1-\underline{\rho}_t),
\end{align*}
which is \eqref{eq:expected-approximation}. 

\subsection{Proof of Corollary~\ref{cor:accuracy-approximation}}
\label{app:proof-accuracy-approximation}

For training accuracy, the best constant-leaf tree has score
$\pi_{\max}$. Algorithm~\ref{alg:candidate-refinement} initializes
the incumbent with this tree and never replaces the incumbent with a
worse solution. Hence,
\[
Q_{A,N}\ge\pi_{\max}.
\]
Since $Q_D^\star\le1$,
\[
\frac{Q_{A,N}}{Q_D^\star}
\ge
Q_{A,N}
\ge
\pi_{\max}.
\]
Thus, the baseline condition in
Proposition~\ref{prop:expected-approximation} holds with
$r_0=\pi_{\max}$. Substitution into
\eqref{eq:expected-approximation} gives
\eqref{eq:accuracy-approximation}.

\subsection{Verification of the Algorithmic Properties}
\label{app:algorithm-properties}

Algorithm~\ref{alg:candidate-refinement} executes at most
$N_{\max}$ iterations because its outer loop is indexed by
$t=1,\ldots,N_{\max}$. Its remaining stopping conditions correspond
to the explicit tests for the runtime limit, maximum score, full
incumbent support, empty proposal block, and patience limit.

The initial candidate set contains at most $K$ random-forest-ranked
features. In every later iteration, CART proposes at most
$K-|C|$ features, and the active set is updated as $E=C\cup N$.
Therefore,
\[
|E|
\le
|C|+|N|
\le
K.
\]
A feasible tree of depth at most $D$ and with at most $M$ internal
nodes uses no more than
\[
\min\{M,2^D-1\}
\]
distinct split features.

For a fixed active set $E$,
Proposition~\ref{prop:objective-equivalence-revised} ensures equivalence between
the weighted and expanded reduced problems whenever its preservation
conditions hold. A certified \texttt{Weighted STreeD} solve therefore
returns an optimum over $\Tset(E,D,M)$. The certificate is specific to
$E$ and does not establish optimality over the full feature set.

Finally, the algorithm replaces the incumbent only when
\[
Q(T;\Iset)>Q^{\mathrm{inc}}+\epsilon.
\]
Otherwise, it retains the existing incumbent. The incumbent-score
sequence is therefore nondecreasing, and an unsuccessful inner solve
does not remove the best feasible tree found previously.

\end{document}